\documentclass[11pt]{article}

\usepackage[final]{acl}

\usepackage{times}
\usepackage{latexsym}
\usepackage{booktabs}
\usepackage{enumitem}

\usepackage[T1]{fontenc}

\usepackage[utf8]{inputenc}

\usepackage{microtype}

\usepackage{inconsolata}

\usepackage{graphicx}

\usepackage{multirow}
\usepackage{amsmath}

\usepackage{algorithm}
\usepackage{algorithmic}

\usepackage{arydshln}

\title{MemVerse: Multimodal Memory for Lifelong Learning Agents}

\author{
  \textbf{Junming Liu}, \textbf{Yifei Sun}, \textbf{Weihua Cheng}, \textbf{Haodong Lei}, \\
  \textbf{Yirong Chen}, \textbf{Licheng Wen}, \textbf{Xuemeng Yang}, \textbf{Daocheng Fu}, \\
  \textbf{Pinlong Cai}, \textbf{Nianchen Deng}, \textbf{Yi Yu}, \textbf{Botian Shi},
  \textbf{Ding Wang}\thanks{Corresponding author.} \\
  Shanghai Artificial Intelligence Laboratory \\
  \texttt{\{liujunming,wangding\}@pjlab.org.cn}
}

\begin{document}

\maketitle

\begin{abstract}
  Despite the remarkable advancements in multimodal large language models across various domains, AI agents still lack the ability to retain past information.
  Without reliable memory, agents struggle to leverage historical context, impairing long-horizon reasoning and coherent interactions in dynamic environments.
  In this paper, we introduce \textbf{MemVerse}, a model-agnostic, plug-and-play memory framework that combines fast parametric recall with hierarchical retrieval, enabling scalable and adaptive multimodal intelligence.
  MemVerse maintains short-term memory for recent context while transforming raw experiences into structured long-term memory graphs, empowering continuous learning and autonomous knowledge evolution.
  To meet real-time demands, we introduce a periodic distillation mechanism that compresses essential knowledge from the cognitive graph into the parametric model, providing fast and differentiable recall while preserving interpretability.
  Extensive experiments show that MemVerse boosts multimodal reasoning with superior efficiency, allowing agents to maintain coherence and adapt seamlessly throughout extended interactions.
Our code is available at \url{https://github.com/KnowledgeXLab/MemVerse}.
\end{abstract}

\begin{figure*}
  \centering
  \includegraphics[width=1.00\textwidth]{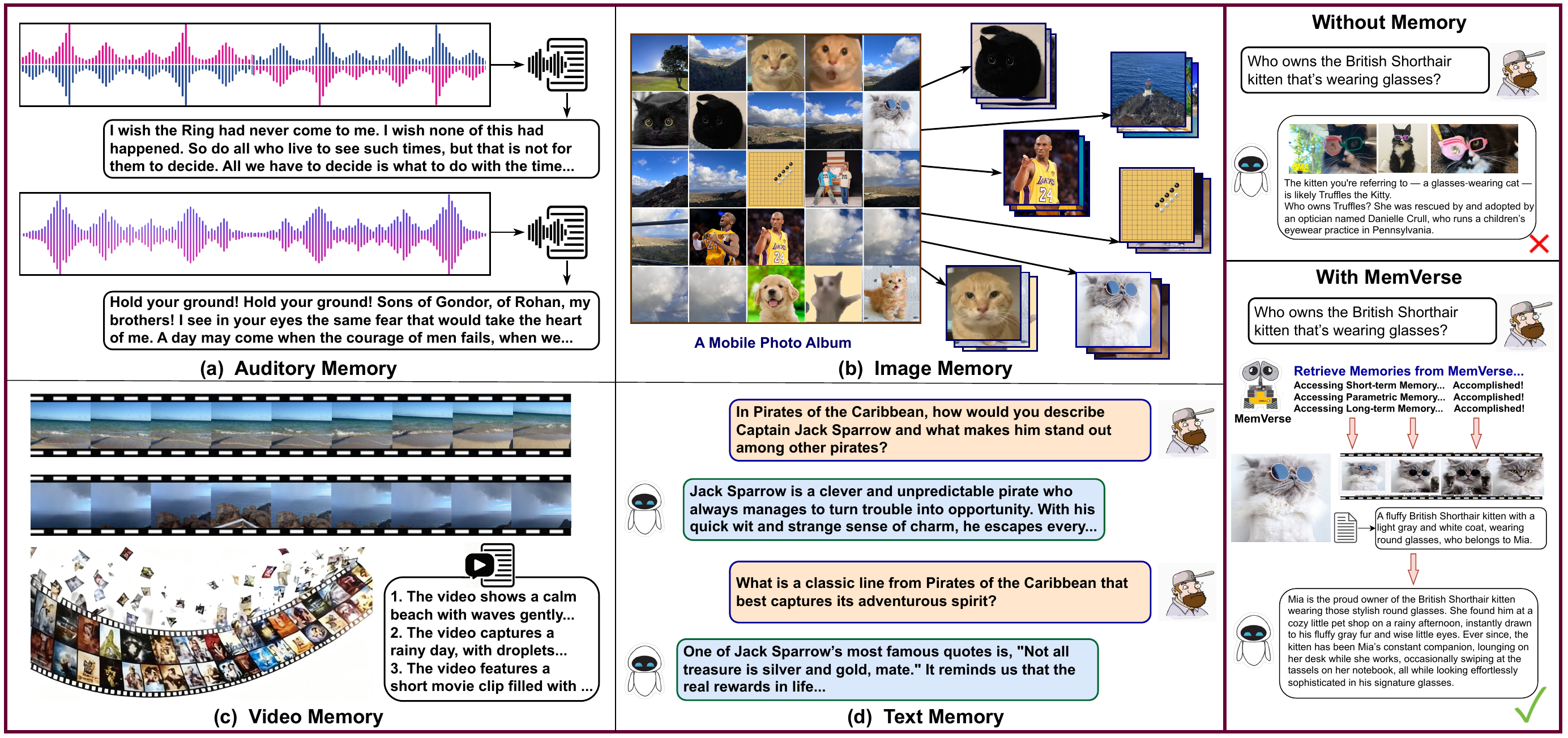}
  \caption{
    (1) The left panels (a-d) illustrate how MemVerse equips agents with multimodal memories such as auditory, visual, video, and textual modalities, enhancing contextual understanding.
    (2) The right panels compare two reasoning outcomes: the baseline LLM without memory (top) tends to hallucinate, while with MemVerse plugged in (bottom), the agent retrieves grounded evidence and generates accurate answers.
  }
  \label{fig:teaser}
\end{figure*}

\section{Introduction}

While large-scale Multimodal Large Language Models (MLLMs) have achieved substantial success \cite{Zhang_2025_NoteLLM2, Wang_2025_AffordBot}, AI agents remain constrained by their lack of effective memory \cite{Zhang_2025_Survey}, particularly in visual reasoning \cite{Dong_2025_Insight-V}, web navigation \cite{Wang_2025_Agent}, and multimodal scene understanding \cite{Kong_2025_Multi}. Memory is essential for enabling agents to adapt to dynamic environments, integrate new experiences with prior knowledge, and perform long-horizon reasoning. Unlike humans, whose intelligence emerges from the continuous accumulation and abstraction of multimodal experiences \cite{Deary2010}, current agents operate in a largely stateless manner \cite{Su_2024_Agents, Liu_2025_VaLiK, Wang_2025_MIRIX}. They often treat each task as an isolated episode, creating a fundamental bottleneck as AI systems shift from single-task inference to continuous interaction in real-world settings.

Current memory solutions fall into two dominant but inadequate paradigms for multimodal \cite{Xiao2025VisualIP, Chhikara_2025_Mem0, Wang_2025_MIRIX} and lifelong learning \cite{Wu_2022_MeMViT, Fan_2024_VideoAgent} scenarios. The first, \textbf{parameter-embedded memory}, encodes knowledge directly into model weights via fine-tuning or in-context learning \cite{pritzel2017nec, Tai_2017_MemNet}. While leveraging the model's capacity for associative recall, it suffers from fixed capacity, costly retraining, catastrophic forgetting, and is inherently a black box that limits interpretability and control. The second relies on \textbf{external storage} via Retrieval-Augmented Generation (RAG) systems \cite{Liu_2025_HM-RAG}, using embedding-based retrieval \cite{Huang_2020_Embedding} or keyword matching \cite{Liu_2006_Effective}. While modular and readily extensible without modifying model parameters, such systems maintain raw interaction logs without abstraction or consolidation, leading to redundancy, retrieval noise, and escalating computational cost as the database grows.

The limitation, however, lies not in either paradigm alone, but in their isolated use. Parameter-embedded memory and external retrieval address different aspects of memory: the former supports compact generalization, whereas the latter preserves explicit and revisitable experiences. This suggests that memory for multimodal agents should not be framed as a choice between parametric and retrieval-based mechanisms, but as a problem of coordinating them. This perspective is consistent with Complementary Learning Systems (CLS) theory~\cite{McClell_1995_CLS, Kahneman2011, evans2013dual}, which holds that human memory achieves both rapid recall and stable generalization through functionally distinct yet complementary systems. The hippocampus preserves episodic detail and relational structure, while the neocortex gradually compresses repeated experiences into efficient representations. Analogously, parametric memory resembles neocortical compression, whereas retrieval memory resembles hippocampal binding: each excels where the other fails. Like human memory, the key is therefore not choosing one over the other, but determining \emph{when} and \emph{how} to rely on each.

Building on this insight, we introduce \textbf{MemVerse} (Figure~\ref{fig:teaser}), a model-agnostic memory framework for AI agents engaged in multimodal reasoning and lifelong learning. MemVerse instantiates the two pathways as two decoupled but jointly governed components: the \textit{slow pathway} operates over hierarchical multimodal knowledge graphs organizing core (user-specific), episodic (time-ordered), and semantic (relational) memory as a dynamic substrate continuously updated from agent experience; the \textit{fast pathway} is a lightweight external parametric memory model, periodically distilled from long-term memory graphs and fully decoupled from the answering LLM, providing low-latency responses.
A central \textit{memory orchestrator} governs routing between the two pathways via a confidence score, making the speed--accuracy trade-off explicit and query-adaptive without learned parameters.
Our key contributions are as follows:
\begin{itemize}[parsep=0pt, itemsep=0pt, leftmargin=1em]
\item We propose MemVerse, a dual-pathway memory architecture that models fast parametric recall and slow graph-based retrieval as complementary, jointly governed components, enabling adaptive routing between them.
\item We design a memory transformation mechanism that converts raw multimodal experiences into specialized memory types organized as hierarchical knowledge graphs, enabling continual abstraction, adaptive forgetting, and bounded memory growth.
\item We introduce a periodic distillation mechanism that compresses essential long-term memory into a lightweight parametric model, enabling fast, differentiable recall while preserving transparency and controllability.
\end{itemize}

\section{Related Work}
\label{sec:relatedwork}

\subsection{Memory for LLM Agents}
Memory is a fundamental component of LLM-based agents~\cite{Zhang_2025_Survey}, with existing approaches spanning parametric and non-parametric paradigms. Parametric memory incorporates knowledge directly into model parameters via trajectory-based fine-tuning~\cite{Chen_2023_Fireact}, modular adaptation~\cite{Yin_2024_AgentLumos}, latent memory tokens~\cite{Wang_2024_MemoryLLM}, or reinforcement-driven optimization~\cite{Yu_2025_MemAgent, Zhang_2025_MemGen}. Non-parametric memory relies on external storage and retrieval: MemGPT~\cite{MemGPT}, MemoryBank~\cite{Zhong_2024_MemoryBank}, and MemoRAG~\cite{Qian_2025_MemoRAG} employ hierarchical retrieval and dual-system mechanisms for long-term reasoning, while SimpleMem~\cite{Liu_2026_SimpleMem}, OCR-Memory~\cite{Li_2026_OCR-Memory}, and Amory~\cite{Zhou_2026_Amory} further improve token efficiency, scalable recall, and narrative coherence. PlugMem~\cite{yang2026plugmem} proposes a task-agnostic plugin memory module for LLM agents. Production systems such as Mem0~\cite{Chhikara_2025_Mem0} and SuperMemory~\cite{Shah_2025_SuperMemory} focus on hierarchical summarization and efficient read/write operations. However, retrieval-based memory suffers from interaction inefficiency, while purely parametric memory lacks flexible long-term retention.

\subsection{Multimodal Knowledge Retrieval}
Multimodal retrieval is central to knowledge-intensive
reasoning in MLLM systems~\cite{Abootorabi_2025_Ask,
Liu_2025_VaLiK}. Recent work explores reward-guided
retrieval~\cite{Fan_2025_End}, position-aware
retrieval~\cite{Yao_2025_Who}, universal multimodal
retrievers~\cite{Lin_2025_MM-EMBED}, synthetic training
pairs~\cite{Zhou_2025_Megapairs}, zero-shot cross-modal
retrieval~\cite{Choi_2025_Zero}, and utility-oriented
evidence selection~\cite{Luo_2026_Utility}.
Knowledge graph-based retrieval has also been explored
for structured reasoning~\cite{zhu2022mmkg, chen2024kgmm,
Wang_2024_Cross}, typically treating graphs as static
external knowledge bases indexed for retrieval.

\begin{figure*}[t]
    \centering
    \includegraphics[width=0.95\linewidth]{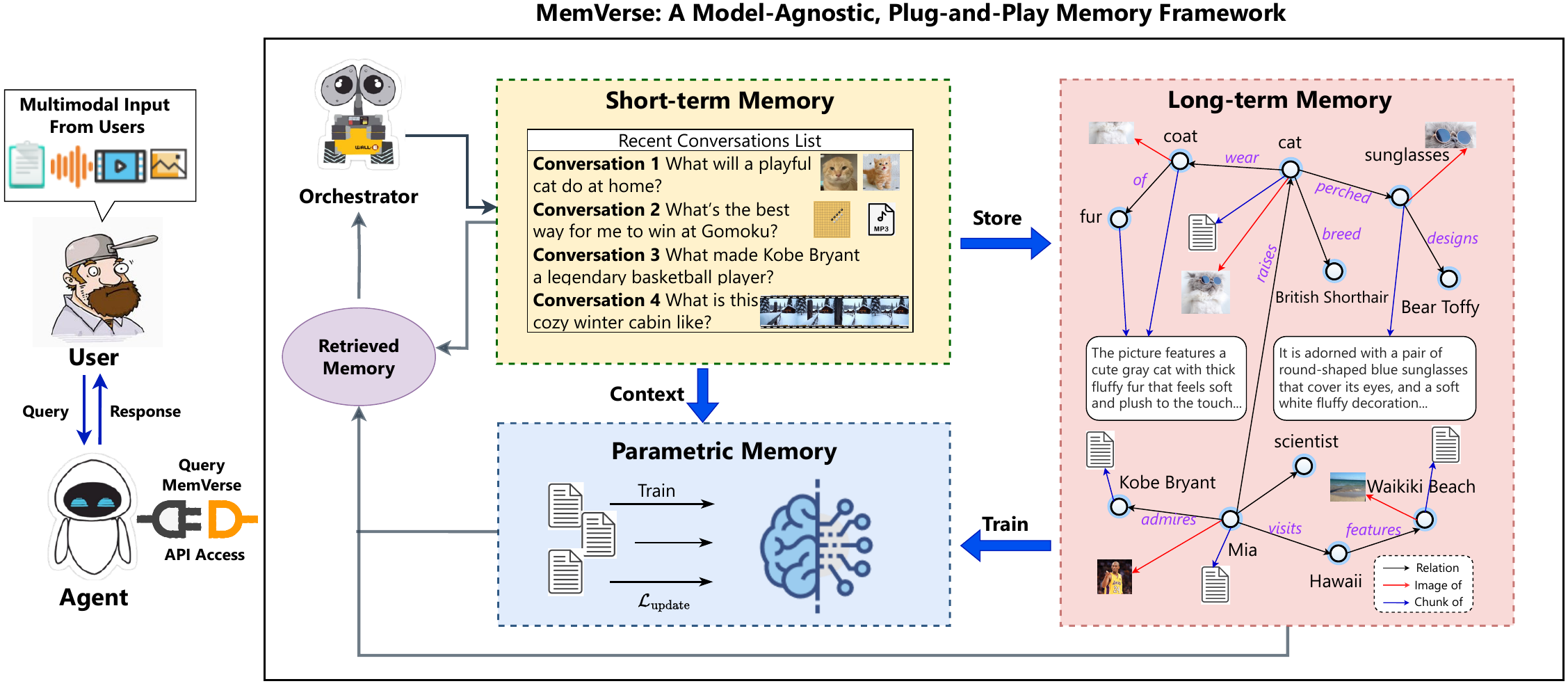}
    \caption{MemVerse integrates three memory components: short-term memory for recent conversational contexts, long-term memory structured as an MMKG with entities and semantic relationships, and parametric memory as a lightweight neural model for fast context encoding. A central memory orchestrator manages retrieval and storage across these components, enabling the agent to process multimodal inputs and support lifelong learning.}
    \label{fig:flow}
\end{figure*}

\section{Methodology}

In this section, we describe the technical design of MemVerse, illustrated in Figure~\ref{fig:flow}. MemVerse adopts a CLS-based memory architecture that combines retrieval memory with parametric memory for long-horizon multimodal reasoning. The framework is centered around a memory orchestrator, which dynamically manages memory addition, update, retrieval, consolidation, and deletion according to the current multimodal context and reasoning state. The retrieval memory contains both short-term and long-term memory. Short-term memory stores recent interaction context, while long-term memory organizes persistent knowledge as MMKGs. Alongside the retrieval memory, the parametric memory provides a fast and differentiable mechanism for knowledge recall and generalization through compact neural representations periodically distilled from long-term memory.

\subsection{Hierarchical Retrieval-based Memory}

\paragraph{Multimodal Processing.}
To handle arbitrary multimodal inputs, such as images, videos, and audio, we first employ pretrained MLLMs to convert the raw data into textual representations. Specifically, each input modality $M$ is encoded into a sequence of descriptive tokens:
\begin{equation}
S = \mathcal{D}_{\text{text}}\Big( \mathcal{A}\big( \mathcal{E}_{\text{mod}}(M) \big) \Big),
\end{equation}
where $M$ denotes the input modality, $\mathcal{E}_{\text{mod}}$ is the modality-specific encoder that extracts features from $M$, $\mathcal{A}$ projects modality-specific representations into a unified textual token space, and $\mathcal{D}_{\text{text}}$ generates the textual tokens. We apply length normalization and redundancy filtering before inserting $S$ into short-term memory.

The resulting text $S$ serves as a text chunk corresponding to the original multimodal input $M$.
This explicit projection into a unified textual token space, together with modality-specific metadata, allows any symbolic entity or relation derived from $S$ to be directly linked back to the original multimodal data \cite{Liu_2025_VaLiK}.
Importantly, \textbf{this design does not introduce irreversible modality information loss}. Although retrieval operates on textual representations, the original multimodal inputs are preserved and can be recovered through the retrieved text chunks and metadata pointers, allowing downstream models to jointly access both the retrieved text and the corresponding raw multimodal data. We adopt textual retrieval because it is generally more efficient and accurate than retrieval in heterogeneous multimodal embedding spaces. Multimodal similarity matching may overlook fine-grained or local information due to interference from dominant modality features \cite{Xie_2025_FG-CLIP, Yao_2025_Who}. For example, a small but task-critical object or character in an image can be missed during multimodal retrieval, while textual representations support both semantic and keyword retrieval, enabling more accurate and controllable access to fine-grained concepts.

\paragraph{Short-Term Memory.}
To avoid redundant retrieval and continuous updates to long-term storage, we introduce a short-term memory mechanism that caches recent interaction states. Specifically, given a dialogue sequence $\{q_1, q_2, ..., q_t\}$, STM retains the most recent $K$ queries within a sliding window:
\begin{equation}
\mathcal{M}_{\text{STM}} = \{ q_{t-K+1}, q_{t-K+2}, ..., q_t \}.
\end{equation}
Since the contextual information in a short dialogue session is already captured within this window, frequent updates to the external memory are unnecessary. Instead, consolidation is performed periodically or when sufficient new knowledge accumulates, ensuring both efficiency and stability.

\paragraph{Long-Term Memory.}
We implement LTM as a paired structure \(\mathcal{M}=(\{\mathcal{G}_k\},\mathcal{C})\), where each \(\mathcal{G}_k=(\mathcal{V}_k,\mathcal{R}_k)\) is a KG corresponding to a specific type of memory (core, episodic, or semantic), and \(\mathcal{C}\) stores the original dialogue text chunks:
{\small
\begin{equation}
\mathcal{M} \;=\; \Big(\; \{\mathcal{G}_k\} \;,\; \mathcal{C} \;\Big),
\quad k \in \{\text{core, episodic, semantic}\}.
\end{equation}
}

Raw dialogue text is first compressed into salient \emph{memory descriptions} by prompting an LLM to retain repeatedly referenced, causally influential, or semantically stable information while discarding transient details. These are then structured into MMKGs. Core memory stores durable user-grounded facts persisting across sessions; episodic memory captures time-stamped, context-dependent interactions; semantic memory maintains user-agnostic abstract knowledge generalizable across users. Entries are routed to subgraphs based on user identity dependency, temporal scope, and abstraction degree.

Formally, given textual chunks \(\mathcal{C}=\{c_i\}\), the LLM extracts entities and typed relations to construct the knowledge graph:
\begin{equation}
\mathcal{G} \;=\; \Phi_{\mathrm{LLM}}(\mathcal{C}) \;=\; (\mathcal{V},\mathcal{R}),
\end{equation}
where each node \(v \in \mathcal{V}\) and each relation \(r \in \mathcal{R}\) maintain persistent references to their supporting text chunks:
{\small
\begin{equation}
\ell_v : \mathcal{V} \to \mathcal{P}(\mathcal{C}), \quad \ell_v(v)=\{\text{Chunks supporting } v\},
\end{equation}
}
{\small
\begin{equation}
\ell_r : \mathcal{R} \to \mathcal{P}(\mathcal{C}), \quad \ell_r(r)=\{\text{Chunks supporting } r\}.
\end{equation}
}

During retrieval, activating an entity or relation in \(\mathcal{G}\) enables access to the corresponding supporting dialogue text \(\mathcal{C}\) and any associated multimodal data \(\mathcal{M}\), ensuring that both symbolic and perceptual knowledge remain accessible and grounded.

\subsection{Lightweight Parametric Memory Model}

Parametric memory is stored as weights in a lightweight language model and updated via periodic supervised fine-tuning on LTM contents. It alleviates the computational overhead of RAG inference by mimicking retrieval behavior and enabling plug-and-play domain adaptation.

\paragraph{Supervised Fine-tuning.}
To enable the model to internalize retrieved knowledge from the LTM, we perform supervised fine-tuning on a constructed dataset of $(q, \mathcal{R}, \hat{\mathcal{R}})$ triplets derived from the LTM retrieval process. Each training instance consists of a question $q$, its multiple-choice context, and the corresponding retrieved answer $\mathcal{R}$ provided by the explicit memory module. The model is optimized to generate $\mathcal{R}$ conditioned on $q$, thereby learning to emulate the retrieval behavior of the LTM within its parametric space.

Formally, the training objective minimizes the token-level cross-entropy loss between the generated sequence $\hat{\mathcal{R}}$ and the target $\mathcal{R}$:
\begin{equation}
\mathcal{L}_{\text{update}}
= - \sum_{t=1}^{T} \log P_{\Theta}(r_t \mid q, r_{<t}),
\end{equation}
where $r_t$ denotes the $t$-th token of the target sequence $\mathcal{R}$, $r_{<t}$ represents its preceding tokens, and $\Theta$ are the current model parameters. This process embeds retrieved knowledge directly into the parameter space, serving as the parametric memory update mechanism.

\paragraph{Dynamic Memory Expansion.}
Parametric memory is updated in coordination with the expansion of the
explicit knowledge graph, serving as a compact and fast-access
complement to retrieval-based memory rather than a sole long-term
knowledge store. At each update step $t$, newly retrieved query–evidence
pairs $(q_t, R_t)$ derived from the explicit memory are used to perform a
lightweight and localized fine-tuning step, yielding an updated
parameter state:
\begin{equation}
\mathcal{M}_{\text{para}}^{t+1} = \mathcal{M}_{\text{para}}^{t} + \Delta \Theta_t,
\end{equation}
where $\Delta \Theta_t$ represents the gradient-based weight update induced by fine-tuning on newly accumulated data. This continual process enables the model to progressively encode new knowledge, ensuring that parametric memory evolves in synchrony with the explicit memory without requiring full retraining.

\section{Experiment}
\subsection{Setups}

\begin{table*}[t]
\centering
\small
\renewcommand{\arraystretch}{0.9}
\setlength{\tabcolsep}{0.8mm}
\begin{tabular}{c|c|ccccc}
\toprule
\multirow{2}{*}{\textbf{Model}} & \multirow{2}{*}{\textbf{Method}} & \multicolumn{5}{c}{\textbf{Answer Prediction (F1)}} \\
& & \textbf{Single Hop} & \textbf{Multi-Hop} & \textbf{Temporal} & \textbf{Open Domain} & \textbf{Overall} \\ \midrule
Human & Human \cite{Maharana_2024_locomo} & 95.1 & 85.5 & 92.6 & 75.4 & 91.6 \\ \midrule
\multirow{8}{*}{Qwen2.5-7B-Instruct}
& Vanilla & 29.3 & \textcolor{cyan}{23.7} & 8.7 & 18.1 & 23.3 \\
& Full Context \cite{epmem} & 10.6 & 3.4 & 3.7 & 8.2 & 7.7 \\
& RAG \cite{epmem} & 21.7 & 8.1 & 4.9 & 14.5 & 15.3 \\
& A-MEM \cite{epmem} & 12.7 & 8.7 & 6.3 & 13.9 & 10.7 \\
& EpMem \cite{epmem} & 17.6 & 6.4 & 10.0 & 2.7 & 13.0 \\
& Mem0$^*$ \cite{Yan_2025_MemoryR1} & 25.0 & 20.3 & \textbf{\textcolor{blue}{33.2}} & \textcolor{cyan}{32.7} & 26.3 \\
& MemoryOS$^*$ \cite{Yan_2025_MemoryR1} & \textcolor{cyan}{29.6} & 21.0 & \textcolor{cyan}{26.3} & \textbf{\textcolor{blue}{40.9}} & \textcolor{cyan}{28.0} \\
& MemVerse & \textbf{\textcolor{blue}{40.3}} & \textbf{\textcolor{blue}{28.1}} & 24.2 & 25.1 & \textbf{\textcolor{blue}{33.8}} \\ \midrule
\multirow{12}{*}{GPT-4o-mini}
& Vanilla & 8.6 & 19.0 & 9.9 & 13.7 & 11.1 \\
& Full Context \cite{A-MEM} & 40.4 & 25.0 & 18.4 & 12.0 & 31.2 \\
& Claude-Mem$^*$ \cite{Liu_2026_Omni-SimpleMem} & 10.2 & 24.5 & 12.2 & 21.5 & 13.9 \\
& A-MEM \cite{A-MEM} & \textcolor{cyan}{44.7} & 27.0 & \textcolor{cyan}{45.9} & 12.1 & \textcolor{cyan}{39.7} \\
& MemGPT \cite{A-MEM} & 41.0 & 26.7 & 25.5 & 9.2 & 33.2 \\
& Zep \cite{Chhikara_2025_Mem0} & 35.7 & 19.4 & 42.0 & \textcolor{cyan}{49.6} & 34.9 \\
& LangMem \cite{Chhikara_2025_Mem0} & 35.5 & 26.0 & 30.8 & 40.9 & 33.1 \\
& Mem0 \cite{Chhikara_2025_Mem0} & 38.7 & 28.6 & \textbf{\textcolor{blue}{48.9}} & 47.7 & 39.6 \\
& MemoryOS \cite{Kang_2025_MemoryOS} & 35.3 & 41.2 & 20.0 & 48.6 & 34.0 \\
& SimpleMem$^*$ \cite{Liu_2026_Omni-SimpleMem} & 12.8 & 30.0 & 17.8 & 31.2 & 18.1 \\
& Omni-SimpleMem$^*$ \cite{Liu_2026_Omni-SimpleMem} & 19.6 & \textbf{\textcolor{blue}{54.4}} & 17.7 & \textbf{\textcolor{blue}{58.8}} & 28.0 \\
& MemVerse & \textbf{\textcolor{blue}{49.2}} & \textcolor{cyan}{46.1} & 29.7 & 30.8 & \textbf{\textcolor{blue}{43.4}} \\ \midrule
\multirow{7}{*}{GPT-5.1}
& Claude-Mem$^*$ \cite{Liu_2026_Omni-SimpleMem} & 17.1 & 28.9 & 26.4 & 29.2 & 22.0 \\
& A-MEM$^*$ \cite{Liu_2026_Omni-SimpleMem} & 16.4 & 28.7 & 24.6 & 28.4 & 21.1 \\
& MemGPT$^*$ \cite{Liu_2026_Omni-SimpleMem} & 16.5 & 28.8 & 24.9 & 29.4 & 21.3 \\
& Mem0$^*$ \cite{Liu_2026_Omni-SimpleMem} & 16.0 & 29.2 & 26.1 & 29.8 & 21.4 \\
& SimpleMem$^*$ \cite{Liu_2026_Omni-SimpleMem} & 17.8 & 30.5 & 27.2 & \textcolor{cyan}{30.5} & 22.9 \\
& Omni-SimpleMem$^*$ \cite{Liu_2026_Omni-SimpleMem} & \textcolor{cyan}{36.7} & \textbf{\textcolor{blue}{59.8}} & \textcolor{cyan}{30.7} & \textbf{\textcolor{blue}{67.6}} & \textcolor{cyan}{41.6} \\
& MemVerse & \textbf{\textcolor{blue}{49.2}} & \textcolor{cyan}{35.6} & \textbf{\textcolor{blue}{38.3}} & 26.7 & \textbf{\textcolor{blue}{43.0}} \\ \midrule
\multirow{3}{*}{GPT-3.5-turbo-16k}
& Vanilla & 34.7 & \textcolor{cyan}{32.5} & 11.0 & 14.4 & 28.1 \\
& RAG \cite{Maharana_2024_locomo} & \textcolor{cyan}{44.3} & 30.6 & \textcolor{cyan}{41.9} & \textbf{\textcolor{blue}{40.2}} & \textcolor{cyan}{41.0} \\
& MemVerse & \textbf{\textcolor{blue}{69.1}} & \textbf{\textcolor{blue}{49.6}} & \textbf{\textcolor{blue}{53.1}} & \textcolor{cyan}{34.2} & \textbf{\textcolor{blue}{60.0}} \\ \bottomrule
\end{tabular}
\begin{flushleft}
\small $^*$  Note: For methods marked with $^*$, we observed discrepancies between the originally reported overall F1 scores and those computed using the official LoCoMo weighted evaluation protocol. We therefore recalculated the overall scores based on the subset distribution: Single-Hop / Multi-Hop / Temporal / Open-Domain = 841 / 282 / 321 / 96.
\end{flushleft}
\caption{Performance comparison on the LoCoMo Benchmark. The Vanilla method represents the evaluation of models without any memory or context enhancement. All other comparative methods utilize data from their respective cited sources. Performance is measured by F1 scores across Single-Hop, Multi-Hop, Temporal, and Open-Domain categories. Higher values denote better performance; the \textbf{\textcolor{blue}{best}} results are highlighted in blue bold, and the \textcolor{cyan}{second-best} results are marked in cyan.}
\label{table:locomo_compare}
\end{table*}

\paragraph{Evaluation Datasets.}
We evaluate the proposed method on four multimodal reasoning and memory benchmarks with diverse characteristics, including long-term conversational memory, science reasoning, and video-text understanding. Specifically, we use \textbf{LoCoMo}~\citep{Maharana_2024_locomo}, a long-horizon multimodal dialogue dataset with extended conversational sessions; \textbf{LongMemEval}~\citep{Wu_2025_LongMemEval}, a large-scale benchmark containing 500 evaluation questions designed for evaluating long-term memory processing capabilities; \textbf{ScienceQA}~\citep{Lu_2022_ScienceQA}, a multimodal science reasoning benchmark; and \textbf{MSR-VTT}~\citep{Xu_2016_MSR-VTT}, a widely used video--text benchmark for cross-modal understanding and retrieval.

\paragraph{Baselines.}
For long-context memory benchmarks including \textbf{LoCoMo} and \textbf{LongMemEval}, we compare against strong proprietary and open-source LLMs, including GPT-3.5-Turbo~\cite{Brown_2020_GPT3}, GPT-4o-mini~\cite{Hurst_2024_Gpt-4o}, Qwen2.5-7B-Instruct~\cite{qwen2025qwen25technicalreport}, and GPT-5.1~\cite{OpenAI_2025_GPT51}.
We further compare with representative memory and retrieval frameworks, including MemoryBank~\cite{Zhong_2024_MemoryBank}, A-MEM~\cite{A-MEM}, EpMem~\cite{epmem}, Mem0~\cite{Chhikara_2025_Mem0}, MemoryOS~\cite{Kang_2025_MemoryOS}, MemGPT~\cite{MemGPT}, Zep~\cite{rasmussen2025zep}, LangMem\footnote{https://langchain-ai.github.io/langmem/}, LD-Agent~\cite{Li_2025_LD-Agent}, QRRetriever~\cite{Zhang_2025_QRRetriever}, SimpleMem~\cite{Liu_2026_SimpleMem}, Omni-SimpleMem~\cite{Liu_2026_Omni-SimpleMem}.
For \textbf{ScienceQA}, we compare against representative text-only and multimodal reasoning models, including CoT~\cite{Lu_2022_ScienceQA}, HoT-T5-Large~\cite{Yao_2023_Thinking}, and Chameleon~\cite{Lu_2023_Chameleon}.
For \textbf{MSR-VTT}, the comparative methods include InternVideo~\cite{wang2022internvideo}, UMT-L~\cite{li2023unmasked}, VAST~\cite{chen2024vast}, Clip4Clip~\cite{luo2022clip4clip}, ExCae~\cite{yang2025expertized}, and CLIP \cite{radford2021clip}.
All models are evaluated using their recommended settings and identical prompting protocols for fair comparison.
Due to space constraints, only representative baselines are discussed here. Descriptions of additional baseline methods, together with implementation setups and optimization details, are provided in Appendix~\ref{details}.

\paragraph{Implementation Details.} For multimodal processing, images, audio, and sampled video frames are converted into textual chunks $\mathcal{C}$ using GPT-4o-mini and Whisper~\cite{Raford_2023_Whisper}, which are then mapped into a hierarchical knowledge graph $\mathcal{G} = (\mathcal{V}, \mathcal{R}) \rightarrow \mathcal{C} \rightarrow M$ spanning core, semantic, and episodic memories~\cite{Liu_2025_VaLiK}. GPT-4o-mini manages both graph construction and retrieval across benchmarks under specialized settings: ScienceQA utilizes its training split for memory and parametric modules; LoCoMo and LongMemEval retain their session-based structures; and MSR-VTT operates in an unsupervised manner without accessing ground-truth text-video alignments. Our default backbone, Qwen2.5-7B, undergoes supervised fine-tuning via a causal language modeling objective using the ``Question-Retrieved'' format. Training is optimized using AdamW with a learning rate of $2 \times 10^{-6}$ over a 2048 sequence length, utilizing linear warm-up, a 1.0 gradient clipping norm, and a single A100 (80G) GPU.

\subsection{Main Results}

\paragraph{LoCoMo}
The results in Table~\ref{table:locomo_compare} demonstrate that MemVerse consistently achieves the state-of-the-art Overall F1 scores across all four evaluated backbones, validating its robustness across varied model scales.
Notably, when paired with GPT-3.5-turbo-16k, MemVerse reaches a peak Overall F1 of \textbf{60.0}, outperforming the RAG baseline by a large margin.
Conversely, the relative improvements on Qwen2.5-7B-Instruct are more modest.
This is primarily attributed to the inherent long-context reasoning constraints of smaller open-source architectures, which face difficulties in effectively prioritizing and integrating dense memory streams.

A task-level breakdown reveals that MemVerse exhibits exceptional efficacy in factual grounding and temporal anchoring, establishing a new state-of-the-art in Single-Hop reasoning across all backbones and achieving strong performance in Temporal tasks.
However, we also observe a performance bottleneck in broader Open Domain and highly complex Multi-Hop scenarios.
A closer error analysis reveals that due to the massive volume and unstructured nature of open-domain history, our dense retrieval mechanism inevitably introduces irrelevant candidate memories.
This semantic noise tends to distract and confuse the downstream generators, particularly when the model attempts to synthesize sparse evidence across multi-step reasoning chains.
To mitigate this limitation, incorporating an explicit semantic reranker \cite{meng2025ranking} to filter out redundant candidates and prune context noise presents a highly promising future direction for further optimizing the framework.

\paragraph{LongMemEval}
As evaluated on the LongMemEval benchmark (Table~\ref{tab:longmemeval_hmo}), MemVerse leverages its graph-based structural topology to achieve exceptionally efficient memory indexing and retrieval, securing a commanding lead with a Rec@5 of \textbf{89.8\%} and an NDCG@5 of \textbf{91.6\%}.
While it delivers a highly competitive downstream accuracy of \textbf{68.4\%}, we observe a distinct performance divergence where the high retrieval recall does not fully translate into proportional answer accuracy.
Our analysis attributes this bottleneck to the \textit{over-reasoning} phenomenon, where the downstream generator over-interprets dense memory inputs and introduces spurious correlations.
To validate this hypothesis, upgrading the downstream reader to the higher-capacity GPT-5.1 drives the final accuracy to a remarkable \textbf{77.0\%}.
This substantial leap demonstrates that the core bottleneck resides in the reasoning capacity of the downstream reader rather than the quality of MemVerse's memory pipeline, proving our framework's capability to supply high-fidelity long-term context.

\begin{table}[t]
\centering
\small
\renewcommand{\arraystretch}{1.1}
\setlength{\tabcolsep}{13pt}
\begin{tabular}{lcc}
\hline
\textbf{Method} & \textbf{Recall@5} $\uparrow$ & \textbf{Accuracy} $\uparrow$ \\ \hline
No History & - & 0.0 \\
Long Context & - & 57.4 \\ \hline
RAG & 62.4 & 63.6 \\ \hline
MemoryBank & 58.6 & 59.6 \\
LD-Agent & 56.8 & 59.2 \\
A-MEM & - & 55.4 \\
Mem0 & - & 65.0 \\
MemoryOS & - & 58.0 \\
QRRetriever & 80.4 & 66.7 \\
\textbf{MemVerse} & \textbf{89.8} & \textbf{68.4} \\ \hline
\end{tabular}
\caption{Comparison of memory method performance on the LongMemEval dataset.}
\label{tab:longmemeval_hmo}
\end{table}

\paragraph{Multimodal Benchmarks.}
To verify the generalizability of MemVerse across diverse multimodal tasks, we evaluate its performance on both the ScienceQA benchmark for multimodal science reasoning and the MSR-VTT dataset for bidirectional video-text retrieval. As shown in Table~\ref{tab:scienceqa} and Table~\ref{tab:msrvtt}, MemVerse consistently enhances foundational backbones, achieving substantial performance gains across both tasks.
On ScienceQA, pairing MemVerse with GPT-4o-mini establishes a new state-of-the-art accuracy of \textbf{85.48\%}, notably outperforming the strong CoT (GPT-4) baseline by 1.49 percentage points and approaching human-level proficiency.

\begin{table}[t]
\centering
\small
\renewcommand{\arraystretch}{1.0}
\setlength{\tabcolsep}{1.6mm}
\begin{tabular}{llcc}
\toprule
\textbf{Category} & \textbf{Method} & \textbf{\#Params} & \textbf{Acc} $\uparrow$ \\ \midrule
\textbf{Baselines} & Human & - & 88.40 \\
& GPT-4 & - & 82.69 \\ \midrule
\textbf{LLM-CoT} & CoT (UnifiedQA) & 223M & 74.11 \\
& CoT (GPT-4) & 1T+ & 83.99 \\
& Chameleon (ChatGPT) & 175B+ & 79.93 \\
& HoT-T5-Large & 738M & 83.38 \\ \midrule
\textbf{MemVerse} & Qwen2.5-7B  & 7B & 75.62 \\
& Qwen2.5-72B & 72B & 80.24 \\
& GPT-4o-mini & - & \textbf{85.48} \\
\bottomrule
\end{tabular}
\caption{Performance on ScienceQA. }
\label{tab:scienceqa}
\end{table}

The MSR-VTT results in Table~\ref{tab:msrvtt} use full-corpus memory.
On the highly competitive MSR-VTT retrieval benchmark, MemVerse demonstrates clear superiority over previous pre-trained foundation models and specialized ViT-based architectures. Specifically, it improves the text-to-video (T2V) \textbf{R@1} to \textbf{90.4\%} and video-to-text (V2T) \textbf{R@1} to \textbf{89.2\%}, surpassing the previous best-performing baseline ExCae by more than $20\%$.
These results demonstrate that MemVerse’s structured memory framework effectively captures cross-modal alignments while supporting complex multi-step reasoning. Due to space constraints, more extensive comparative experiments and detailed task-level statistics are provided in Appendix~\ref{more_results}.

\begin{table}[t]
\centering
\small
\renewcommand{\arraystretch}{1.0}
\setlength{\tabcolsep}{1.4mm}
\begin{tabular}{llcccc}
\toprule
\multirow{2}{*}{\textbf{Cat.}} & \multirow{2}{*}{\textbf{Method}} & \multicolumn{2}{c}{\textbf{T2V}} & \multicolumn{2}{c}{\textbf{V2T}} \\
\cmidrule(lr){3-4} \cmidrule(lr){5-6}
& & \textbf{R@1} & \textbf{R@5} & \textbf{R@1} & \textbf{R@5} \\ \midrule
\textbf{Pre-trained} & InternVideo & 55.2 & 79.6 & 57.9 & 79.2 \\
& UMT-L & 58.8 & 81.0 & 58.6 & 81.6 \\
& VAST & 63.9 & 84.3 & - & - \\ \midrule
\textbf{ViT} & Clip4Clip & 46.4 & 72.1 & 45.4 & 73.4 \\
& ExCae & 67.7 & 92.7 & 69.3 & 92.5 \\
& CLIP & 29.7 & 48.9 & 21.4 & 38.6 \\ \midrule
\textbf{MemVerse} & GPT-4o-mini & \textbf{90.4} & \textbf{95.6} & \textbf{89.2} & \textbf{92.7} \\
\bottomrule
\end{tabular}
\caption{Performance on MSR-VTT Retrieval. }
\label{tab:msrvtt}
\end{table}

\subsection{Ablation Study}

Table~\ref{tab:ablation} presents the ablation results on the LoCoMo benchmark. Vanilla models and those with only a sliding window (\textit{+Short-Term}) underperform due to limited context, whereas independently integrating the parametric module (\textit{+Param}) or the graph-based retrieval system (\textit{+Long-Term}) substantially enhances results; for instance, \textit{+Long-Term} improves the Overall F1 from $11.1$ to $39.9$ on GPT-4o-mini and from $23.3$ to $32.1$ on Qwen2.5-7B-Instruct. Combining both memory streams via our routing strategy (\textit{+Orchestrator}) yields the optimal overall performance, reaching \textbf{43.4} and \textbf{33.8} F1 scores respectively. Although the orchestrator introduces slight trade-offs in Multi-Hop and Temporal tasks under the Qwen backbone compared to pure \textit{+Long-Term} retrieval, the consistent improvements in overall scores validate that our hybrid memory approach effectively captures long-range dependencies beyond standard context windows.

\begin{table}[t]
\centering
\small
\renewcommand{\arraystretch}{1.0}
\setlength{\tabcolsep}{1.2mm}
\begin{tabular}{l|ccccc}
\toprule
\textbf{Method} & \textbf{S-Hop} & \textbf{M-Hop} & \textbf{Temp.} & \textbf{Open} & \textbf{Over.} \\ \midrule
\textit{GPT-4o-mini} & & & & & \\
Vanilla & 8.6 & 19.0 & 9.9 & 13.7 & 11.1 \\
+Short-Term & 11.1 & 17.7 & 8.9 & 11.7 & 11.9 \\
+Param & 48.9 & 28.7 & 19.9 & 11.7 & 36.8 \\
+Long-Term & 47.0 & 40.8 & 25.9 & 22.5 & 39.9 \\
MemVerse & 49.2 & 46.1 & 29.7 & 30.8  & 43.4 \\ \midrule
\textit{Qwen2.5-7B-Ins.} & & & & & \\
Vanilla & 29.3 & 23.7 & 8.7 & 18.1 & 23.3 \\
+Short-Term & 27.2  & 24.5 & 10.2 & 16.8 & 22.5 \\
+Param & 28.8 & 24.2 & 10.1 & 19.2 & 23.5 \\
+Long-Term & 36.9 & 29.1 & 24.8 & 22.7 & 32.1 \\
MemVerse & 40.3 & 28.1 & 24.2 & 25.1 & 33.8 \\
\bottomrule
\end{tabular}
\caption{Ablation study on the LoCoMo dataset.}
\label{tab:ablation}
\end{table}

\subsection{Further Analysis}
\label{further_analysis}

\begin{table}[t]
\centering
\small
\renewcommand{\arraystretch}{1.2}
\setlength{\tabcolsep}{1.5mm}
\begin{tabular}{lccccc}
\hline
\textbf{Method} & \textbf{S-Hop} & \textbf{M-Hop} & \textbf{Temp.} & \textbf{Open} & \textbf{Over.} \\ \hline
Flat RAG & 30.3 & 27.3 & 18.5 & 19.3 & 26.6 \\
MMKG-only & \textbf{47.0} & \textbf{40.8} & \textbf{25.9} & \textbf{22.5} & \textbf{39.9} \\ \hline
\end{tabular}
\caption{Comparison between Flat RAG and MMKG-only under GPT-4o-mini.}
\label{tab:mmkg_vs_rag}
\end{table}

\paragraph{Benefits of Structured Graph Memory.}
To isolate the advantages of structured memory, we compare a Flat RAG vector store baseline against an MMKG-only configuration using GPT-4o-mini, as shown in Table~\ref{tab:mmkg_vs_rag}. Compared with flat retrieval, MMKG-only achieves consistent improvements across all task categories, with particularly large gains in Multi-Hop and Temporal reasoning. These results demonstrate that hierarchical graph memory structures are significantly more effective than flat text retrieval for modeling complex dependencies.

\begin{table}[t]
\centering
\small
\renewcommand{\arraystretch}{1.1}
\setlength{\tabcolsep}{8pt}
\begin{tabular}{lll}
\hline
\textbf{Component} & \textbf{Metric} & \textbf{Value} \\ \hline
Memory Extraction & Total Time & 22.03 s \\
& Per 1k Tokens & 0.20 s \\ \hline

MMKG Construction & Total Time & 208.56 s \\
& Per Chunk & 13.03 s \\
& Per 1k Tokens & 16.94 s \\ \hline

Parametric Training & Total Time & 978.72 s \\
& Training Tokens & 453,841 \\
& Per 1k Tokens & 2.16 s \\ \hline
\end{tabular}
\caption{Efficiency analysis of memory extraction, MMKG construction, and parametric memory distillation on the LoCoMo benchmark.}
\label{tab:parametric_efficiency}
\end{table}

\paragraph{Efficiency of Parametric Distillation.}
We provide a detailed breakdown of the computational cost for memory extraction, MMKG construction, and parametric memory distillation in Table~\ref{tab:parametric_efficiency}. The results show that the overall pipeline remains computationally efficient, with memory extraction requiring only $0.20$s per 1k tokens and parametric memory training scaling at $2.16$s per 1k training tokens. Although the total training time is relatively longer, this is primarily due to multiple training epochs and the large amount of contextual training data, rather than inefficiency in the memory module itself.

\paragraph{Knowledge Retention and Scalability.}
We investigate the periodic update strategy of parametric memory on the ScienceQA dataset using a grouped partitioning scheme to examine knowledge retention across intervals, with details in Appendix~\ref{Evolution}. Furthermore, we evaluate MemVerse's scalability by scaling the backbone from Qwen2.5-1.5B to 72B while keeping the Qwen2.5-7B-based memory module fixed, isolating the impact of model capacity on reasoning accuracy as discussed in Appendix~\ref{Scalability}.

\begin{table*}[t]
\centering
\small
\begin{tabular}{llrrrr}
\toprule
\textbf{Memory access} & \textbf{Method} & \textbf{T2V R@1} & \textbf{T2V R@5} & \textbf{V2T R@1} & \textbf{V2T R@5} \\
\midrule
\multirow{4}{*}{Full corpus} & A-Mem & 68.8 & 80.4 & 67.3 & 78.5 \\
& Mem0 & 77.4 & 86.6 & 78.6 & 90.2 \\
& SimpleMem & 53.5 & 72.0 & 56.7 & 81.5 \\
& MemVerse & \textbf{90.4} & \textbf{95.6} & \textbf{89.2} & \textbf{92.7} \\
\midrule
\multirow{2}{*}{No full-corpus access} & CLIP & 29.7 & 48.9 & 21.4 & 38.6 \\
& MemVerse (train-only) & \textbf{45.2} & \textbf{76.2} & \textbf{45.3} & \textbf{76.9} \\
\bottomrule
\end{tabular}
\caption{MSR-VTT retrieval results (\%) under two memory access protocols. CLIP is the reference without full-corpus memory access.}
\label{tab:msrvtt_memory_protocols}
\end{table*}

\begin{figure*}[t]
\centering
\includegraphics[width=\textwidth]{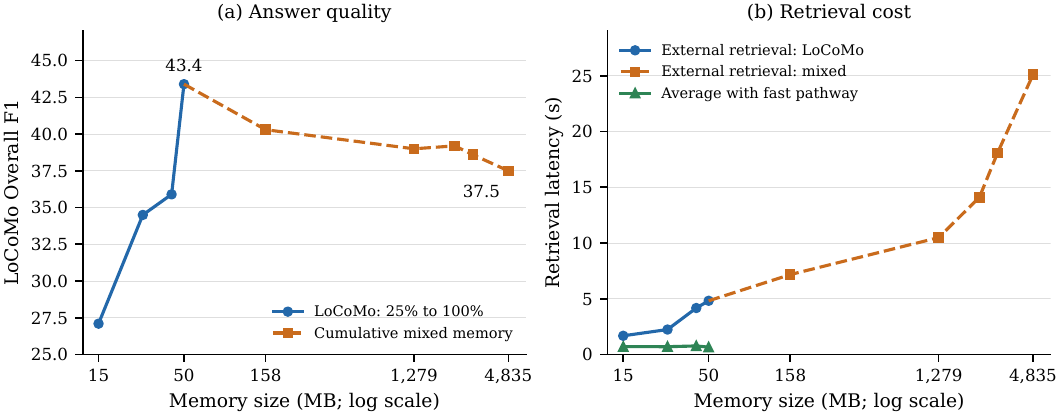}
\caption{Memory growth on LoCoMo with GPT-4o-mini. Blue points show 25\%, 50\%, 75\%, and 100\% in-domain memory (15, 28, 42, and 50 MB). Orange points show cumulative mixed-memory expansion (158, 1,279, 2,270, 2,930, and 4,835 MB). The right panel distinguishes external-memory retrieval latency from average latency with the fast pathway enabled.}
\label{fig:memory_growth}
\end{figure*}

\paragraph{Memory Access Protocols.}
We evaluate full-corpus and train-only memory access on MSR-VTT in Table~\ref{tab:msrvtt_memory_protocols}. Under full-corpus access, all memory systems receive the same video corpus and shuffled corpus captions without video--caption alignment labels. MemVerse achieves 90.4\% T2V R@1 and 89.2\% V2T R@1. With train-only memory, MemVerse reaches 45.2\% and 45.3\%, respectively.

\paragraph{Memory Growth.}
We evaluate LoCoMo with GPT-4o-mini under a fixed test set while growing in-domain memory from 15 MB to 50 MB, then cumulatively adding memories from ScienceQA, MSR-VTT, LongMemEval, MemoryAgentBench~\cite{Hu_2026_MemoryAgentBench}, and MedMemoryBench~\cite{Wang_2026_MedMemoryBench}. Figure~\ref{fig:memory_growth} shows Overall F1 increasing from 27.1 to 43.4 with in-domain memory and reaching 37.5 in the largest mixed pool (4,835 MB). External retrieval latency increases from 4.82 s at full LoCoMo memory to 25.11 s in the largest mixed pool. Parametric memory is not retrained for the mixed pools; average latency with the fast pathway is reported only for in-domain pools.

\section{Conclusion}

We introduced MemVerse, a model-agnostic, plug-and-play memory framework that brings AI agents closer to lifelong multimodal intelligence. MemVerse unifies fast parametric recall with slow, hierarchical retrieval-based memory, mirroring the complementary roles of intuition and deliberation. This dual-path design lets agents maintain recent context, transform raw multimodal experiences into structured knowledge, and retrieve information efficiently without unbounded memory growth.
Across diverse multimodal tasks, MemVerse yields significant improvements in reasoning accuracy, stability, and long-horizon coherence. These results demonstrate that scalable memory, rather than ever-larger models, is the missing ingredient for continuous-learning agents. Looking ahead, we plan to explore more adaptive memory-control strategies and to deploy MemVerse in open-world environments across a variety of domains, fostering agents that learn, remember, and truly evolve.

\section{Limitations}

Although MemVerse demonstrates strong effectiveness on long-context memory and multimodal reasoning benchmarks, its current graph-based memory organization has so far been evaluated primarily under small- to medium-scale memory settings, as discussed in Section~\ref{further_analysis}. When deployed in longer-term interactions, the memory graph may continue to expand, increasing retrieval and insertion overhead and potentially introducing noise from outdated memories. A promising direction is to dynamically partition memories into multiple temporal graphs, where new graphs are periodically initialized to prevent excessive node accumulation while preserving retrieval efficiency. Historical memories can then be accessed through temporal indexing over archived graphs when needed.

\section*{Ethical Considerations}

Our experiments use existing public benchmarks. Persistent memory can retain personal information and reproduce biased or inaccurate content. Applications involving user data should obtain user consent, restrict access to stored memories, and support correction and deletion.

\section*{Acknowledgements}

\paragraph{AI Use.}
We used Codex to assist with manuscript revision, reference verification, and the preparation of figures and supporting scripts. The authors are responsible for the final content.

\bibliography{memverse}

\appendix

\newpage
\clearpage

\section{Datasets}
\label{sec:dataset}

\paragraph{LoCoMo~\citep{Maharana_2024_locomo}.}
LoCoMo\footnote{https://github.com/snap-research/locomo} is a benchmark for long-term multimodal dialogue understanding. It consists of \textbf{10 high-quality dialogues}, each spanning up to \textbf{32 sessions}, with an average of \textbf{600 turns} and \textbf{16,618 tokens} per conversation. The dialogues are generated by LLM-based agents guided by \textit{unique personas} and \textit{temporal event graphs}, resulting in coherent and contextually rich conversations over long horizons. Each dialogue supports \textbf{image sharing} and \textbf{reaction annotations}, providing a combination of textual and visual modalities. LoCoMo emphasizes temporal reasoning, persona consistency, and multimodal grounding over extended interactions, making it a particularly demanding benchmark for models that must maintain context and interpret visual cues over hundreds of turns.

\paragraph{LongMemEval~\citep{Wu_2025_LongMemEval}.}
LongMemEval is a benchmark proposed by Wu et al.~\citep{Wu_2025_LongMemEval} for evaluating long-term memory capabilities in LLM-based conversational assistants. It is designed to assess whether models can effectively store, retrieve, update, and reason over information across long and multi-session interactions.
The benchmark evaluates several key memory abilities, including:
\begin{itemize}
    \item \textbf{Information Extraction}: extracting salient facts from historical conversations;
    \item \textbf{Multi-session Reasoning}: reasoning across multiple dialogue sessions;
    \item \textbf{Temporal Reasoning}: understanding temporal relations and state changes;
    \item \textbf{Knowledge Updates}: handling user information updates and overwriting outdated facts;
    \item \textbf{Abstention}: refusing to answer when the required information is absent.
\end{itemize}
LongMemEval contains hundreds of manually curated questions embedded in realistic multi-session chat histories with very long contexts, making it substantially more challenging than traditional short-context benchmarks. The benchmark also studies memory system designs such as indexing, retrieval, and memory-aware query expansion for improving long-context conversational QA performance.
Experimental results show that even state-of-the-art long-context LLMs and commercial conversational assistants exhibit significant performance degradation in long-term memory scenarios, highlighting the limitations of current memory mechanisms in persistent conversational settings.

\paragraph{ScienceQA~\citep{Lu_2022_ScienceQA}.}
ScienceQA\footnote{https://github.com/lupantech/ScienceQA} is a multimodal dataset designed for science question answering, containing \textbf{21,208 questions} across natural science, social science, and language science. Approximately \textbf{48.7\% of instances include images}, such as diagrams, charts, and illustrations, which require models to perform \textit{cross-modal reasoning} between textual descriptions and visual content. Each question is multiple-choice with two to five candidate answers. In addition to images, ScienceQA provides \textbf{image captions} to facilitate reasoning for text-only models, allowing fair comparisons between unimodal and multimodal approaches. The dataset is particularly challenging because visual content often conveys critical information that is not explicitly described in the question text, necessitating integrated visual–textual comprehension.

\paragraph{MSR-VTT~\citep{Xu_2016_MSR-VTT}.}
MSR-VTT\footnote{https://www.microsoft.com/en-us/research/publication/msr-vtt-large-video-description-dataset-bridging-video-language-supplementary-material/} is a large-scale video–text dataset widely used for video understanding and cross-modal retrieval. It contains \textbf{10,000 YouTube video clips} spanning \textbf{20 diverse categories} such as sports, music, and cooking. Each video is paired with \textbf{20 human-annotated captions}, resulting in \textbf{200K video–sentence pairs}. Videos vary in duration, visual complexity, and motion patterns, posing challenges for models in \textit{temporal alignment}, \textit{action recognition}, and \textit{semantic grounding} of dynamic visual content. MSR-VTT is commonly used to benchmark \textit{video–text retrieval}, \textit{captioning}, and \textit{cross-modal reasoning} capabilities, requiring models to integrate spatiotemporal cues from videos with natural-language semantics.

Collectively, these datasets cover diverse multimodal reasoning scenarios: LoCoMo and LongMemEval evaluate \textit{long-term conversational context and persona reasoning}, ScienceQA emphasizes \textit{scientific diagram comprehension}, and MSR-VTT tests \textit{video–text alignment and dynamic visual understanding}. They provide a comprehensive evaluation suite for models capable of integrating visual and textual information across different temporal and semantic scales.

\begin{figure}[t]
    \centering
    \begin{minipage}{0.45\textwidth}
        \centering
        \includegraphics[width=\linewidth]{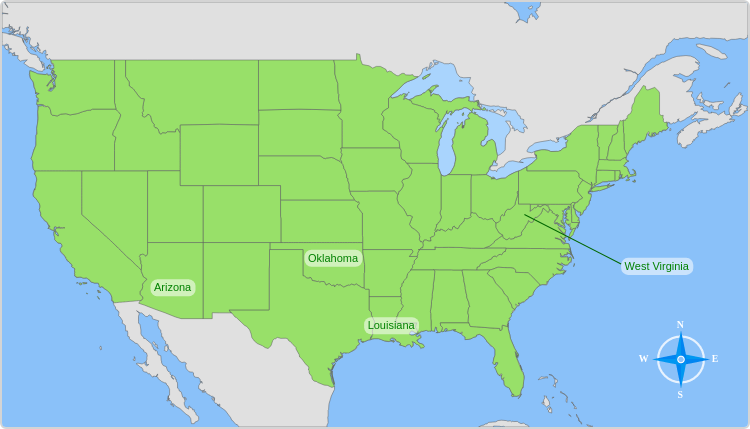}
    \end{minipage}

    \vspace{0.8em}

    \begin{minipage}{0.45\textwidth}
        \small
        \textbf{Question:} Which of these states is farthest north?\\[0.3em]
        \textbf{Choices:}
        \begin{itemize}
            \item West Virginia
            \item Louisiana
            \item Arizona
            \item Oklahoma
        \end{itemize}
    \end{minipage}
    \caption{An example instance from the ScienceQA dataset, showing a question with a corresponding diagram.}
    \label{fig:scienceqa_example}
\end{figure}

\begin{figure}[htbp]
    \centering
    \begin{minipage}{0.45\textwidth}
        \centering
        \includegraphics[width=\linewidth]{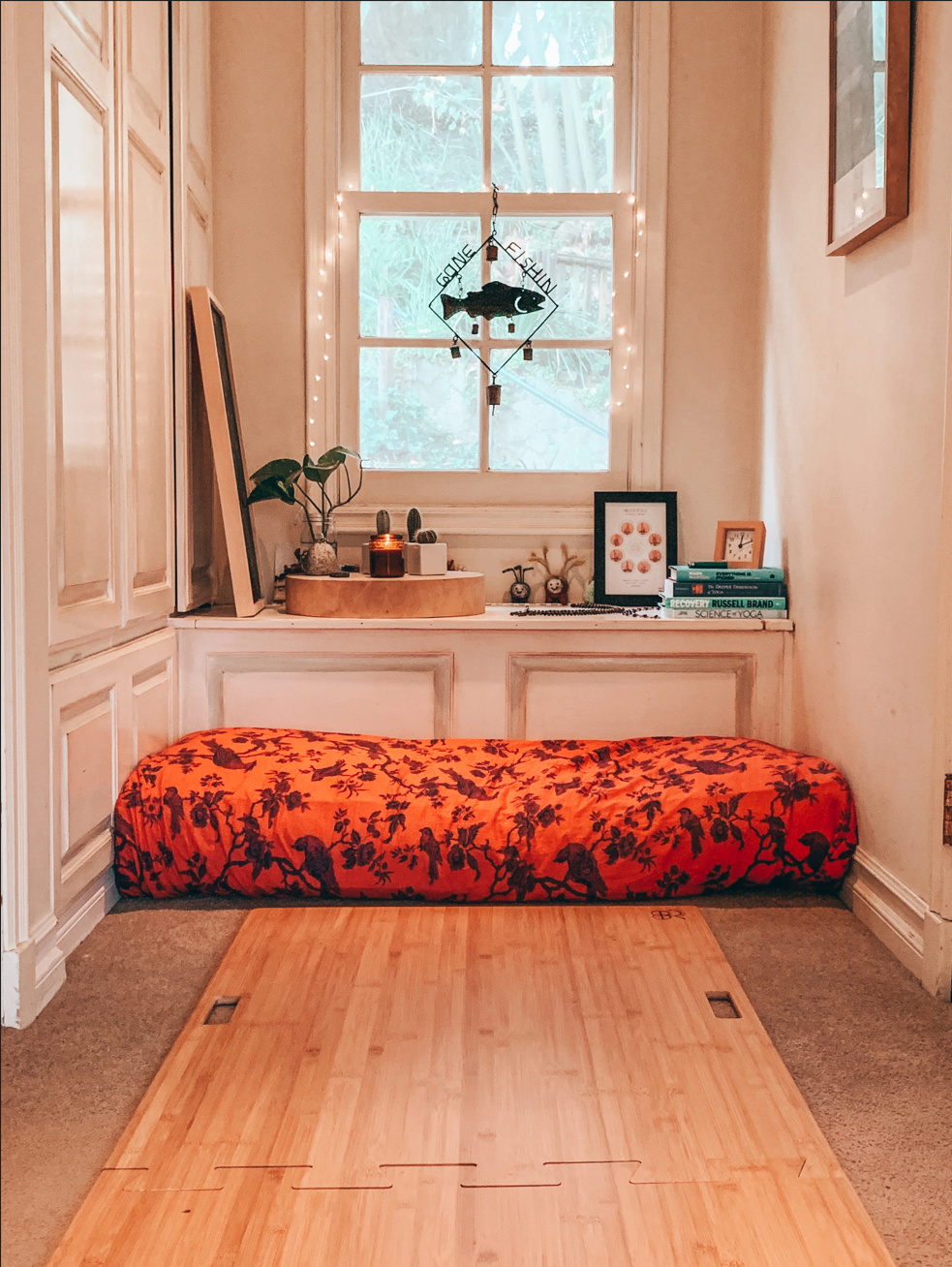}
    \end{minipage}

    \vspace{0.8em}

    \begin{minipage}{0.45\textwidth}
        \small
        \textbf{Speaker:} Jolene\\
        \textbf{BLIP Caption:} a photo of a room with a bench and a window\\
        \textbf{Query:} serene yoga studio windows natural light\\
        \textbf{Dialogue ID:} D1:6\\[0.3em]
        \textbf{Text:}\\
        Sorry about your loss, Deb. My mother also passed away last year.
        This is my room in her house; I also have many memories there.
        Is there anything special about it you remember?
    \end{minipage}

    \caption{A representative visual-text dialogue snippet from the LoCoMo dataset. The figure shows the shared image along with its associated caption, speaker metadata, query, and conversational text.}
    \label{fig:locomo_example}
\end{figure}

\begin{figure}[htbp]
    \centering
    \includegraphics[width=0.45\textwidth]{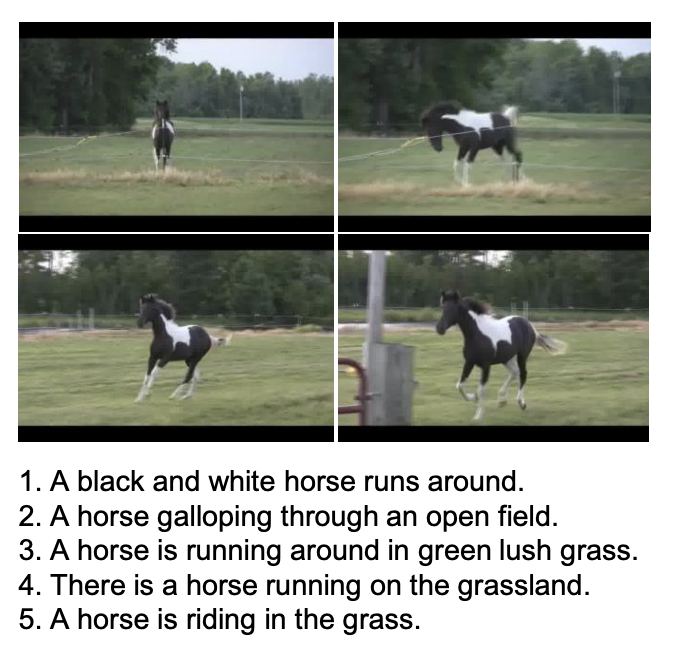}
    \caption{An example frame from the MSR-VTT dataset, paired with its corresponding caption.}
    \label{fig:msrvtt_example}
\end{figure}

Example instances from each dataset are shown in Figures~\ref{fig:scienceqa_example}--\ref{fig:msrvtt_example}, illustrating the typical content and modalities of LoCoMo, ScienceQA, and MSR-VTT.

\section{Implementation Details}
\label{details}

\subsection{Baselines}

For long-context memory benchmarks including \textbf{LoCoMo} and \textbf{LongMemEval}, we compare against strong proprietary and open-source LLMs, including GPT-3.5-Turbo~\cite{Brown_2020_GPT3}, GPT-4o-mini~\cite{Hurst_2024_Gpt-4o}, Qwen2.5-7B-Instruct~\cite{qwen2025qwen25technicalreport}, and GPT-5.1~\cite{OpenAI_2025_GPT51}.
We further compare with representative memory and retrieval frameworks, including MemoryBank~\cite{Zhong_2024_MemoryBank}, A-MEM~\cite{A-MEM}, EpMem~\cite{epmem}, Mem0~\cite{Chhikara_2025_Mem0}, MemoryOS~\cite{Kang_2025_MemoryOS}, MemGPT~\cite{MemGPT}, Zep~\cite{rasmussen2025zep}, LangMem\footnote{https://langchain-ai.github.io/langmem/}, LD-Agent~\cite{Li_2025_LD-Agent}, QRRetriever~\cite{Zhang_2025_QRRetriever}, SimpleMem~\cite{Liu_2026_SimpleMem}, Omni-SimpleMem~\cite{Liu_2026_Omni-SimpleMem}.
For \textbf{ScienceQA}, we compare models across general-domain reasoning in zero- and few-shot settings, covering both text-only and multimodal paradigms.
Specifically, text-based LLMs include GPT Model~\cite{Lu_2022_ScienceQA}, CoT~\cite{Lu_2022_ScienceQA}, HoT-T5-Large \cite{Yao_2023_Thinking} and DDCoT~\cite{Zheng_2023_Advances}, while multimodal baselines include LaVIN~\cite{Luo_2023_LaVIN}, BLIP-2, CCOT~\cite{Mitra_2024_CCOT}, and GraphVis~\cite{Deng_2024_GraphVis}.
We also include the tool-augmented VLM Chameleon~\cite{Lu_2023_Chameleon} and the RAG-enhanced model VaLiK~\cite{Liu_2025_VaLiK} as strong multimodal reasoning baselines.
For \textbf{LoCoMo}, we include GPT-3.5-Turbo~\cite{Brown_2020_GPT3}, GPT-4o-mini \cite{Hurst_2024_Gpt-4o}, and Qwen2.5-7B-Instruct~\cite{qwen2025qwen25technicalreport}, covering strong proprietary and open-source models with robust reasoning abilities. We conduct a comprehensive comparison against several state-of-the-art memory-augmented frameworks, including A-MEM \cite{A-MEM}, EpMem \cite{epmem}, Mem0 \cite{Chhikara_2025_Mem0}, MemoryOS \cite{Kang_2025_MemoryOS}, MemGPT \cite{MemGPT}, Zep \cite{rasmussen2025zep}, and LangMem\footnote{https://langchain-ai.github.io/langmem/}. All models are tested under identical prompting and evaluation protocols to ensure fair cross-modal comparison.
For \textbf{MSR-VTT}, We compare our approach with a comprehensive set of representative methods, including InternVideo \cite{wang2022internvideo}, UMT-L \cite{li2023unmasked}, CLIP-VIP \cite{xue2022clip}, mPLUG-2 \cite{xu2023mplug}, VAST \cite{chen2024vast}, CLIP2TV \cite{gao2021clip2tv}, DRL \cite{wang2022disentangled}, TS2-Net \cite{liu2022ts2}, CLIP4Clip \cite{luo2022clip4clip}, X-CLIP \cite{ma2022x}, DMAE \cite{jiang2023dual}, Cap4Video++ \cite{cap4video++}, TeachClip \cite{holistic}, ExCae \cite{yang2025expertized} and the CLIP \cite{radford2021clip} baseline. Our baseline uses only the CLIP text encoder, which yields an extremely lightweight model compared with video based architectures.

\subsection{Multimodal Preprocessing}

For all experiments, we convert multimodal inputs, including images, audio, and videos, into unified textual representations. Images are captioned using GPT-4o-mini \cite{Hurst_2024_Gpt-4o}; audio is transcribed via Whisper \cite{Raford_2023_Whisper}; videos are either uniformly sampled into frames and captioned with a VLM. Each resulting text segment is treated as an independent \textit{chunk}.

Following prior work on multimodal knowledge grounding \cite{Liu_2025_VaLiK}, we additionally enrich each chunk using the CoE principle \cite{Xiao_2024_CoE}, which injects contextual, observational, and explanatory information where appropriate. This enriches the downstream memory extraction and improves knowledge completeness.

\subsection{Memory Extraction Prompts}

We use a unified prompting pipeline to convert the textualized multimodal chunks, together with the original textual data, into three forms of memory: core, semantic, and episodic. The templates used for LLM extraction shown in Tables~\ref{tab:core_memory_prompt}--\ref{tab:episodic_memory_prompt}.
These extracted memories constitute the parameterized memory corpus used during downstream retrieval and reasoning.

\subsection{Knowledge Graph Construction}

We then apply an LLM-based entity and relation extraction step to convert each memory into a structured MMKG \cite{Liu_2025_VaLiK}. Among several existing frameworks, we draw inspiration from LightRAG \cite{Guo_2024_LightRAG} due to its robustness and scalability in similar settings. Accordingly, we adopt a graph construction strategy akin to theirs, creating one graph per memory instance. This decoupled design improves scalability: graph insertion remains efficient as each graph stays relatively small, and retrieval avoids the bottlenecks typically observed in monolithic, large-scale knowledge graphs.

\subsection{Training Details}

We build MMKGs from the training split of ScienceQA, the session-level dialogues of LoCoMo, and the full MSR-VTT dataset (excluding its labels), all of which serve as the basis for parameterized memory generation. Qwen2.5-7B is used as the backbone language model for supervised fine-tuning. The objective is to encode the multimodal knowledge distilled from ScienceQA, LoCoMo, and MSR-VTT directly into the model parameters.

Training follows the causal language modeling paradigm. Each training example is constructed as a \textbf{Question-Retrieved} pair, and the loss is applied only to the output segment. Key hyperparameters include: sequence length of 2048, AdamW optimizer with a learning rate of $2 \times 10^{-6}$, linear learning rate scheduler with 10\% warm-up, and gradient clipping with a maximum norm of 1.0. All experiments are conducted on a single A100~80GB GPU.

\section{Algorithms}

The core procedures of our memory management system are detailed in the following algorithms. Specifically, Algorithm \ref{alg:memverse_query} outlines the online memory inference workflow, while the subsequent memory update and construction process is formalized in Algorithm \ref{alg:memverse_construct}.

\begin{algorithm}[t]
\caption{MemVerse Online Memory Inference}
\label{alg:memverse_query}
\begin{algorithmic}[1]

\STATE \textbf{Input:} query $q$, short-term memory $\mathcal{M}_{\text{STM}}$, long-term memory $\{\mathcal{G}_{\text{core}}, \mathcal{G}_{\text{episodic}}, \mathcal{G}_{\text{semantic}}\}$, parametric memory $\mathcal{M}_{\text{para}}$
\STATE \textbf{Output:} response $\hat{y}$ and memory context $\mathcal{R}$

\STATE Initialize $\mathcal{R} \leftarrow \emptyset$
\STATE $\mathcal{R} \leftarrow \mathcal{R} \cup \mathcal{M}_{\text{STM}}$

\STATE $\mathcal{R}_{\text{para}} \leftarrow \text{Retrieve}(q, \mathcal{M}_{\text{para}})$
\STATE $\mathcal{b} \leftarrow \text{Orchestrator}(q, \mathcal{R}_{\text{para}})$

\IF{$\mathcal{b} = \text{true}$}

    \STATE $\mathcal{R} \leftarrow \mathcal{R} \cup \mathcal{R}_{\text{para}}$
    \STATE $\hat{y} \leftarrow f_{\text{LLM}}(q, \mathcal{R})$
    \STATE \textbf{return} $\hat{y}, \mathcal{R}$

\ENDIF

\STATE $\mathcal{R}_{\text{LTM}} \leftarrow \emptyset$

\FOR{each $\mathcal{G}_k \in \{\mathcal{G}_{\text{core}}, \mathcal{G}_{\text{episodic}}, \mathcal{G}_{\text{semantic}}\}$}
    \STATE $\mathcal{Z}_1^{k} \leftarrow \text{TopK}(q, \mathcal{G}_k)$ \COMMENT{1-hop seed entities}
    \STATE $\mathcal{Z}_2^{k} \leftarrow \text{Neighborhood}(\mathcal{Z}_1^{k}, \mathcal{G}_k)$ \COMMENT{2-hop expansion}
    \STATE $\mathcal{R}_{\text{LTM}} \leftarrow \mathcal{R}_{\text{LTM}} \cup \mathcal{Z}_1^{k} \cup \mathcal{Z}_2^{k}$
\ENDFOR

\STATE $\mathcal{R} \leftarrow \mathcal{R} \cup \mathcal{R}_{\text{LTM}}$

\STATE $\hat{y} \leftarrow f_{\text{LLM}}(q, \mathcal{R})$

\STATE \textbf{return} $\hat{y}, \mathcal{R}$

\end{algorithmic}
\end{algorithm}

\begin{algorithm}[t]
\caption{MemVerse Memory Construction and Parametric Distillation}
\label{alg:memverse_construct}
\begin{algorithmic}[1]

\STATE \textbf{Input:} conversation history $\mathcal{C}=\{(q_t, a_t)\}$, long-term memory $\{\mathcal{G}_{\text{core}}, \mathcal{G}_{\text{episodic}}, \mathcal{G}_{\text{semantic}}\}$, orchestrator $\mathcal{O}$, parametric memory $\mathcal{M}_{\text{para}}$
\STATE \textbf{Output:} updated memory structures $\{\mathcal{G}_k\}$ and $\mathcal{M}_{\text{para}}$

\STATE $(\mathcal{M}_{\text{core}},
\mathcal{M}_{\text{episodic}},
\mathcal{M}_{\text{semantic}})
\leftarrow \mathcal{O}_{\text{extract}}(\mathcal{C})$
\COMMENT{extract and categorize memory entries}

\FOR{each memory type $k \in \{\text{core}, \text{episodic}, \text{semantic}\}$}

    \FOR{each memory entry $m \in \mathcal{M}_{k}$}

        \STATE $(v_i, r_{ij}, v_j)
        \leftarrow \mathcal{O}_{\text{parse}}(m)$
        \COMMENT{extract entities and relations}

        \STATE $(v_i', v_j')
        \leftarrow \mathcal{O}_{\text{merge}}(v_i, v_j, \mathcal{G}_k)$
        \COMMENT{entity deduplication and alignment}

        \STATE $\mathcal{G}_k
        \leftarrow \mathcal{O}_{\text{update}}
        (\mathcal{G}_k, v_i', r_{ij}, v_j')$
        \COMMENT{relation insertion and graph update}

    \ENDFOR

\ENDFOR

\IF{$t \bmod T = 0$}

    \STATE $\mathcal{X}_{\text{para}}
    \leftarrow \mathcal{O}_{\text{distill}}(\{\mathcal{G}_k\})$
    \COMMENT{construct training pairs from LTM}

    \STATE $\mathcal{M}_{\text{para}}
    \leftarrow \text{SFT}
    (\mathcal{M}_{\text{para}}, \mathcal{X}_{\text{para}})$

\ENDIF

\STATE \textbf{return} $\{\mathcal{G}_k\}, \mathcal{M}_{\text{para}}$

\end{algorithmic}
\end{algorithm}

\section{Additional Results}
\label{more_results}

\begin{table*}[t!]
    \centering
    \renewcommand{\arraystretch}{0.85}
    \small
	\resizebox{\textwidth}{!}{
		\begin{tabular}{lcccccccccc}
			\toprule
			\multirow{2}{*}{\textbf{Method}} &\multirow{2}{*}{\textbf{\#Params}} & \multicolumn{3}{c}{\textbf{Subject}} & \multicolumn{3}{c}{\textbf{Context Modality}} & \multicolumn{2}{c}{\textbf{Grade}} &  \multirow{2}{*}{\textbf{Average}} \\
			&& \textbf{NAT}   & \textbf{SOC}   & \textbf{LAN}   & \textbf{TXT}   & \textbf{IMG}   & \textbf{NO}    & \textbf{G1-6}  & \textbf{G7-12} &    \\
			\midrule
			Human~\cite{Lu_2022_ScienceQA} &-&  90.23 & 84.97 & 87.48 & 89.60 & 87.50 & 88.10 & 91.59 & 82.42 & 88.40 \\ \midrule
			GPT-4~\cite{Liu_2023_LLaVa} &-& 84.06 & 73.45 & 87.36 & 81.87 & 70.75 & 90.73 & 84.69 & 79.10 & 82.69 \\
                CoT (GPT-3)~\cite{Lu_2022_ScienceQA} &173B&  75.44 & 70.87 & 78.09 & 74.68 & 67.43 & 79.93 & 78.23 & 69.68 & 75.17 \\
			CoT (UnifiedQA)~\cite{Lu_2022_ScienceQA} &223M& {71.00} &  76.04 &  {78.91} &  {66.42} &  {66.53} & {81.81} &  {77.06} & 68.82 &  {74.11} \\
			CoT (GPT-4)~\cite{Lu_2023_Chameleon} &1T+& \textcolor{blue}{85.48} &  {72.44} &  \textcolor{blue}{90.27} &  82.65 &  {71.49} & \textcolor{blue}{92.89} &  \textcolor{cyan}{86.66} & 79.04 &  \textcolor{cyan}{83.99} \\
			DDCoT~\cite{Zheng_2023_Advances} &175B & 80.15 & 76.72 & 82.82 & 78.89 & 72.53 & 85.02 & 82.86 & 75.21 & 80.15 \\
            HoT-T5-Large \cite{Yao_2023_Thinking} &738M& 84.46 &  79.08 &  84.64 &  \textcolor{cyan}{82.89} &  \textcolor{cyan}{75.81} & 88.15 &  83.88 & \textcolor{blue}{82.47} &  83.38 \\
                Chameleon (ChatGPT)~\cite{Lu_2023_Chameleon} &175B+& 81.62 &  70.64 &  84.00 &  79.77 &  70.80 & 86.62 &  81.86 & 76.53 &  79.93 \\  \midrule
			LaVIN-13B~\cite{Yang_2023_Mm} &13B & -	&- &-&	-&-	&- &	-& -& 77.54\\
                BLIP-2~\cite{Yang_2023_Mm} &- & -	&- &-&	-&-	&- &	-& -& 74.17\\
                CCOT~\cite{Mitra_2024_CCOT} &7B & -	&- &-&	-&-	&- &	-& -& 76.84\\
                GraphVis~\cite{Deng_2024_GraphVis} &7B & -	&- &-&	-&-	&- &	-& -& 73.18\\
			\midrule
			Qwen2.5-7B~\cite{Liu_2025_VaLiK} & 7B & 76.20 &  67.83  &  77.27  & 74.49 &  65.79 &  79.02 &  77.72 &  69.35 & 74.72 \\
                Qwen2.5-7B (Mmkg)~\cite{Liu_2025_VaLiK} & 7B & 73.98 &  66.37  &  78.18  & 71.65 &  64.30 &  79.65 &  76.51 &  68.03 & 73.47 \\
                Qwen2.5-7B (Visual Genome)~\cite{Liu_2025_VaLiK}& 7B & 76.78 &  67.04  &  78.09  & 74.05 &  66.19 &  79.72 &  78.08 &  69.68 & 75.08 \\
                Qwen2.5-7B (MemVerse) & 7B & 74.51 &  74.58 &  78.73 & 75.92 &  66.19 &  81.95 &  79.70 &  68.29 & 75.62 \\
                Qwen2.5-72B~\cite{Liu_2025_VaLiK} & 72B & 79.64 &  67.10  &  84.90  & 77.56 &  65.00 &  87.93 &  80.25 &  74.85 & 78.37 \\
                Qwen2.5-72B (MemVerse) & 72B & 77.53 & \textcolor{cyan}{80.76} &  85.36  & 78.68 &  66.39 &  89.20 &  82.31 &  76.53 & 80.24 \\
                GPT-4o-mini & - & 77.31 &  63.10  &  86.91  & 74.05 & 66.86 &  87.93 &  83.37 &  65.06 & 76.82 \\
                GPT-4o-mini (MemVerse) & - & \textcolor{cyan}{85.26} &  \textcolor{blue}{81.55} &  \textcolor{cyan}{89.09} & \textcolor{blue}{83.28} &  \textcolor{blue}{78.19} &  \textcolor{cyan}{91.50} &  \textcolor{blue}{88.11} &  \textcolor{cyan}{80.75} & \textcolor{blue}{85.48} \\
			\bottomrule
		\end{tabular}
	}
	\caption{Performance comparison (\%) on ScienceQA benchmark. \#Params denotes backbone parameter counts. Categories: NAT (natural science), SOC (social science), LAN (language), TXT (text context), IMG-Cap (image caption), NO (no context), G1-6 (grades 1-6), G7-12 (grades 7-12).
    Method groups: (1) Human performance baseline, (2) Zero/Few-shot text-only LLMs, (3) Zero/Few-shot Multimodal VLMs, (4) LLMs/VLMs enhanced with memory mechanisms or external knowledge for multimodal reasoning. The best results are highlighted in \textbf{\textcolor{blue}{blue}}, while the second-best results are denoted in \textcolor{cyan}{light blue}.
	}
		\label{tab:main}
\end{table*}

\begin{table*}[t]
    \centering
    \small
    \setlength{\tabcolsep}{3mm}
    \renewcommand{\arraystretch}{1.0}
    \begin{tabular}{lc|ccc|ccc}
        \toprule
        \multirow{2}{*}{\textbf{Method}} & \multirow{2}{*}{\textbf{Backbone}} & \multicolumn{3}{c|}{\textbf{Text-to-Video}} & \multicolumn{3}{c}{\textbf{Video-to-Text}} \\
        \cline{3-8}
          & & R@1$\uparrow$ & R@5$\uparrow$ & R@10$\uparrow$ & R@1$\uparrow$ & R@5$\uparrow$ & R@10$\uparrow$ \\
        \midrule
        \multicolumn{8}{l}{ \textit{\textbf{Pre-trained Foundation Model}} } \\
        InternVideo \cite{wang2022internvideo}  & \textit{ViT-H/14} & 55.2 & 79.6 & 87.5 & 57.9 & 79.2 & 86.4 \\
        UMT-L \cite{li2023unmasked}  & \textit{ViT-L/14} & 58.8 & 81.0 & 87.1 & 58.6 & 81.6 & 86.5 \\
        CLIP-VIP \cite{xue2022clip} & \textit{ViT-B/16} & 57.7 & 80.5 & 88.2 & -- & -- & -- \\
        mPLUG-2 \cite{xu2023mplug} & \textit{ViT-L/14} & 53.1 & 77.6 & 84.7 & -- & -- & -- \\
        VAST \cite{chen2024vast} & \textit{ViT-G/14} & 63.9 & 84.3 & 89.6 & -- & -- & -- \\
        \midrule
        \multicolumn{8}{l}{ \textit{\textbf{ViT-based}} } \\
        CLIP2TV \cite{gao2021clip2tv} & \textit{ViT-B/16} & 49.3 & 74.7 & 83.6 & 46.9 & 75.0 & 85.1 \\
        DRL \cite{wang2022disentangled} & \textit{ViT-B/16} & 49.4 & 76.4 & 84.2 & 47.0 & 77.1 & 84.4 \\
        TS2-Net \cite{liu2022ts2} & \textit{ViT-B/16} & 47.8 & 76.8 & 85.2 & 47.8 & 76.0 & 84.6 \\
        Clip4Clip \cite{luo2022clip4clip} & \textit{ViT-B/16} & 46.4 & 72.1 & 82.0 & 45.4 & 73.4 & 82.4 \\
        X-CLIP \cite{ma2022x} & \textit{ViT-B/16} & 49.3 & 75.8 & 84.8 & 48.9 & 76.8 & 84.5 \\
        DMAE \cite{jiang2023dual} & \textit{ViT-B/16} & 49.9 & 75.8 & 85.5 & 49.6 & 76.3 & 85.0 \\
        Cap4Video++ \cite{cap4video++} & \textit{ViT-B/16} & 52.3 & 76.8 & 85.8 & 50.0 & 75.9 & 86.0 \\
        TeachClip \cite{holistic} & \textit{ViT-B/16} & 48.0 & 75.9 & 83.5 & -- & -- & -- \\
        ExCae \cite{yang2025expertized} & \textit{ViT-G/14} & 67.7 & 92.7 & 96.2 & 69.3 & 92.5 & 96.3 \\
        CLIP \cite{radford2021clip} & \multicolumn{1}{c}{-} & 29.7 & 48.9 & 58.8 & 21.4 & 38.6 & 44.3 \\
        MemVerse & \multicolumn{1}{c}{-} & \textbf{\textcolor{blue}{90.4}} & \textbf{\textcolor{blue}{95.6}} & \textbf{\textcolor{blue}{98.1}} & \textbf{\textcolor{blue}{89.2}} & \textbf{\textcolor{blue}{92.7}} & \textbf{\textcolor{blue}{99.0}} \\
        \bottomrule
    \end{tabular}
    \caption{Comparison with SOTA methods on the MSR-VTT dataset. The \textbf{\textcolor{blue}{best}} results are highlighted in blue bold.}
    \label{tab:msr_comparison}
\end{table*}

\paragraph{ScienceQA}
Table~\ref{tab:main} presents the performance of various methods on the ScienceQA benchmark.
Overall, our \textbf{MemVerse} enhanced models achieve the state-of-the-art performance across most evaluation metrics. Specifically, the highest accuracy in terms of average score is \textbf{85.48\%}, achieved by the GPT-4o-mini equipped with MemVerse.
For subject-specific scores, the MemVerse-enhanced model also obtains strong performance in \textit{natural science} with 85.26\%, \textit{social science} with 81.55\%, and \textit{language} with 89.09\%.
In terms of context modalities, it achieves the best results for both \textit{text context} with 83.28\% and \textit{image caption} with 78.19\%, demonstrating that our memory mechanism effectively leverages multimodal knowledge.

It is worth noting that, on the ScienceQA dataset, \textit{short-term memory} contributes relatively little, as the test questions are largely non-sequential with limited contextual dependencies.
The \textit{parametric memory}, trained using the \textit{long-term memory}, achieves comparable accuracy to long-term retrieval but with significantly faster access.
Specifically, using the RAG approach requires on average 20.17 seconds per question, while retrieving from the compressed long-term memory only takes 8.26 seconds on average.
The parametric memory further reduces the average retrieval time to 2.28 seconds, reducing retrieval latency by approximately 89\% compared to RAG and 72\% compared to long-term retrieval, while maintaining similar performance.
This demonstrates that parametric memory offers a practical trade-off between speed and effectiveness in knowledge-augmented reasoning.

\paragraph{MSR-VTT}
To evaluate generalization, we conduct experiments on the MSR-VTT dataset. During inference, the query text retrieves relevant entries from the memory graph to perform query rewriting before final matching. As shown in Table~\ref{tab:msr_comparison}, \textbf{MemVerse} achieves R@1 scores of \textbf{90.4\%} (text-to-video) and \textbf{89.2\%} (video-to-text), outperforming the CLIP baseline by \textbf{60.7} and \textbf{67.8} percentage points, respectively. This dramatic leap demonstrates that memory-augmented retrieval significantly enhances semantic matching.

Notably, our approach achieves these improvements without exposing the ground-truth alignment between captions and videos to the knowledge graph or the parametric module, ensuring fair evaluation. During memory construction, pairs of captions are partially aligned and connected through LLM's powerful understanding, judgment, and reasoning capabilities, forming linked representations that are stored in the memory.
During retrieval, the query text can effectively leverage these stored associations, resulting in highly accurate matching.
As a result, this memory-based approach overcomes the limitations of directly using large LLMs or VLMs for retrieval over massive multimodal candidate pools. Training such large video models is extremely costly due to the scale of the data, and their limited context windows prevent them from processing all candidates simultaneously. By storing semantic relationships extracted with the understanding, judgment, and reasoning capabilities of models like GPT-4o-mini, our method allows lightweight embedding-based models to efficiently retrieve highly relevant entries while still leveraging the rich reasoning captured in the memory. This effectively combines the scalability of embedding-based retrieval with the semantic power of large pretrained models, demonstrating the key advantage of our approach.

\section{Memory Evolution Performance}
\label{Evolution}

Table \ref{tab:increase} analyzes how the parametric memory evolves as the model undergoes periodic updates with incrementally accumulated memory.  We simulate a realistic memory growth scenario in which the model internalizes retrieved knowledge in sequential stages. The baseline corresponds to the original Qwen2.5-7B model without any parametric memory, while the subsequent rows (\(25\%\), \(50\%\), \(75\%\), \(100\%\)) represent the model after fine-tuning on the earliest \(k\%\) segments of the retrieved memory. This setting allows us to examine how partial, temporally ordered memory updates influence long-term retention, stability, and generalization.

The results indicate that early memory updates already provide meaningful improvements in several modality categories. With only \(25\%\) of accumulated knowledge, the model exhibits stronger parametric recall in retrieval-oriented modalities such as NO and IMG, suggesting that even early retrieved examples offer strong supervisory signals for internalizing memory access patterns. Increasing the update interval to \(50\%\) yields broader gains across subject areas---particularly SOC and NAT---demonstrating that mid-stage memory provides more semantically diverse supervision that enhances the stability of the learned memory.

The most favorable performance emerges at the 75\% update stage, achieving the highest overall accuracy (\textbf{75.69}). The observed plateau and slight degradation in performance when using 100\% of the memory can be attributed to the nature of the ScienceQA dataset and the strength of the Qwen2.5-7B base models. ScienceQA primarily tests broadly applicable scientific knowledge, which the base model already captures quite well; as a result, adding more memory provides little additional benefit. In contrast, for smaller models such as 1.5B, the memory is far from sufficient, and increasing the memory size yields more noticeable improvements. These observations highlight the importance of periodic, well-paced memory updates rather than relying solely on full-memory consolidation, as the benefits depend on both model capacity and dataset characteristics.
Overall, these results validate the design motivation of our parametric memory mechanism: incremental updates enable the model to gradually absorb retrieved knowledge while preserving generalization.

\section{Scalability Analysis of MemVerse}
\label{Scalability}

\begin{table*}[t!]
\centering
\setlength{\tabcolsep}{2.5mm}
\begin{tabular}{cccccccccc}
\hline
\multicolumn{1}{c}{\multirow{2}{*}{\textbf{Method}}} & \multicolumn{3}{c}{\textbf{Subject}}                                                                   & \multicolumn{3}{c}{\textbf{Context Modality}}                                                          & \multicolumn{2}{c}{\textbf{Grade}}                                     & \multicolumn{1}{c}{\multirow{2}{*}{\textbf{Average}}} \\
\multicolumn{1}{c}{}                                 & \multicolumn{1}{c}{\textbf{NAT}} & \multicolumn{1}{c}{\textbf{SOC}} & \multicolumn{1}{c}{\textbf{LAN}} & \multicolumn{1}{c}{\textbf{TXT}} & \multicolumn{1}{c}{\textbf{IMG}} & \multicolumn{1}{c}{\textbf{NO}} & \multicolumn{1}{c}{\textbf{G1-6}} & \multicolumn{1}{c}{\textbf{G7-12}} & \multicolumn{1}{c}{}                                  \\ \hline
Qwen2.5-7B & 76.20 & 67.83 & 77.27 & 74.49 & 65.79 & 79.02 & 77.72 & \underline{69.35} & 74.72 \\
+25\% Memory & 76.07 & 68.95 & 78.14 & 74.33 & \underline{65.94} & \underline{83.34} & 79.11 & 67.96 & 75.12 \\
+50\% Memory & \underline{76.49} & \underline{69.85} & 77.82 & 74.65 & 65.49 & 81.74 & 78.56 & \textbf{69.81} & 75.43 \\
+75\% Memory & \textbf{76.89} & 66.48 & \textbf{80.64} & \underline{75.07} & 65.81 & \textbf{85.71} & \textbf{80.43} & 67.17 & \textbf{75.69} \\
+100\% Memory & 74.51 & \textbf{74.58} & \underline{78.73} & \textbf{75.92} & \textbf{66.19} & 81.95 & \underline{79.70} & 68.29 & \underline{75.62} \\ \hline
\end{tabular}
\caption{Performance of the parametric memory under periodic knowledge accumulation.
The baseline corresponds to the original Qwen2.5-7B model without memory updates, while the 25\%, 50\%, 75\%, and 100\% rows represent staged fine-tuning with progressively larger partitions of retrieved knowledge.
This setup simulates incremental memory growth and reveals how intermediate update intervals (50\%--75\%) yield the most stable and effective parametric memory consolidation.}
\label{tab:increase}
\end{table*}

\begin{table*}[t]
\centering
\setlength{\tabcolsep}{2mm}
\begin{tabular}{cccccccccc}
\hline
\multirow{2}{*}{\textbf{Method}} & \multicolumn{3}{c}{\textbf{Subject}}             & \multicolumn{3}{c}{\textbf{Context Modality}}     & \multicolumn{2}{c}{\textbf{Grade}} & \multirow{2}{*}{\textbf{Average}} \\
                                 & \textbf{NAT}   & \textbf{SOC}   & \textbf{LAN}   & \textbf{TXT}   & \textbf{IMG}   & \textbf{NO}    & \textbf{G1-6}    & \textbf{G7-12}  &                                   \\ \hline
Qwen2.5-1.5B                     & -              & -              & -              & -              & -              & -              & -                & -               & 53.93                             \\
+Param.                 & -              & -              & -              & -              & -              & -              & -                & -               & \textbf{61.54}                    \\
Qwen2.5-7B & \textbf{76.20} & 67.83 & 77.27 & 74.49 & 65.79 & 79.02 & 77.72 & \textbf{69.35} & 74.72 \\
+Param. & 74.51 & \textbf{74.58} & \textbf{78.73} & \textbf{75.92} & \textbf{66.19} & \textbf{81.95} & \textbf{79.70} & 68.29 & \textbf{75.62} \\
Qwen2.5-14B & 75.67 & 67.83 & 81.91 & 70.58 & 63.96 & 85.99 & 79.99 & 67.83 & 75.64 \\
+Param. & \textbf{75.98} & \textbf{72.44} & \textbf{82.36} & \textbf{71.26} & \textbf{65.25} & \textbf{86.76} & \textbf{80.69} & \textbf{70.01} & \textbf{76.87} \\
Qwen2.5-32B & 77.53 & 69.29 & 85.00 & 72.68 & \textbf{66.58} & 88.78 & 82.34 & 69.48 & 77.74 \\
+Param. & \textbf{78.11} & 69.29 & \textbf{87.36} & \textbf{72.92} & 66.39 & \textbf{90.87} & \textbf{82.71} & \textbf{71.39} & \textbf{78.66} \\
Qwen2.5-72B & \textbf{79.64} & 67.10 & 84.90 & 77.56 & 65.00 & 87.93 & 80.25 & 74.85 & 78.37 \\
+Param. & 77.53 & \textbf{80.76} & \textbf{85.36} & \textbf{78.68} & \textbf{66.39} & \textbf{89.20} & \textbf{82.31} & \textbf{76.53} & \textbf{80.24} \\ \hline
\end{tabular}
\caption{Scalability Analysis of parametric Memory on ScienceQA Benchmark (\%) Categories: NAT (natural science), SOC (social science), LAN (language), TXT (text context), IMG-Cap (image caption), NO (no context), G1-6 (grades 1-6), G7-12 (grades 7-12).
   Method groups: (1) Qwen2.5 models with varying scales (1.5B, 7B, 14B, 32B, 72B) as base models; (2) Corresponding base models enhanced with Parametric Memory (fixed Qwen2.5-7B parametric memory module). For each model scale, \textbf{bold} indicates the better-performing result between the base model and its parametric memory-enhanced variant.
	}
		\label{tab:scale}
\end{table*}

Table~\ref{tab:scale} presents a comprehensive comparison of baseline Qwen2.5 models and their counterparts augmented with parametric memory across multiple subject categories, context modalities, and grade levels. A consistent trend emerges: incorporating parametric memory leads to noticeable performance improvements across nearly all model scales, demonstrating its effectiveness as a lightweight enhancement mechanism.

For smaller models such as Qwen2.5--1.5B, parametric memory yields a substantial gain of $\mathbf{+7.61}$ points on the overall average, indicating that limited-capacity models benefit significantly from externalized abstraction and compressed history integration. As model size increases, the magnitude of improvement becomes more nuanced. For mid-sized models (7B and 14B), parametric memory continues to provide stable gains of $\mathbf{+0.90}$ and $\mathbf{+1.23}$ points, respectively, suggesting that even moderately capable models profit from additional memory-based supervision for stabilizing reasoning across heterogeneous subject and modality conditions.

Larger models exhibit an even more interesting behavior. For Qwen2.5--32B, parametric memory contributes a solid improvement of $\mathbf{+0.92}$ points, particularly boosting performance in categories such as natural science (NAT), language (LAN), and reasoning-heavy modalities (NO). Notably, the 72B model, despite already strong baseline performance, still benefits by $\mathbf{+1.87}$ points, showing that parametric memory remains complementary to large-scale pretraining. The gains on social science (SOC) questions are especially pronounced, indicating that parametric memory enhances multi-step reasoning and cross-modal grounding.

Overall, the results demonstrate that parametric memory consistently improves accuracy across scales, with the largest relative benefits observed in smaller models and meaningful enhancements retained even at the 72B level. This highlights the robustness and scalability of the proposed parametric memory mechanism, illustrating its role as an efficient tool for strengthening knowledge retention and improving generalization across diverse tasks.

\section{More Related Work Discussion}

\subsection{Memory for LLM Agents.}

Memory is a fundamental component of LLM-based agents, shaping their ability to adapt, generalize, and operate over long-horizon interactions~\cite{Zhang_2025_Survey}. Existing approaches can generally be categorized into parametric and non-parametric paradigms. Parametric memory incorporates knowledge directly into model parameters through mechanisms such as trajectory-based fine-tuning~\cite{Chen_2023_Fireact}, modular adaptation~\cite{Yin_2024_AgentLumos}, or latent memory tokens~\cite{Wang_2024_MemoryLLM}. More recent methods further combine parametric learning with latent or reinforcement-driven memory optimization, including MemAgent~\cite{Yu_2025_MemAgent} and MemGen~\cite{Zhang_2025_MemGen}.
Non-parametric memory instead relies on external storage and retrieval systems to maintain persistent and scalable context. Representative works such as MemGPT~\cite{MemGPT}, MemoryBank~\cite{Zhong_2024_MemoryBank}, and MemoRAG~\cite{Qian_2025_MemoRAG} employ hierarchical retrieval, temporal relevance modeling, and dual-system memory mechanisms for long-term reasoning. Recent studies further emphasize memory efficiency and structured consolidation. SimpleMem~\cite{Liu_2026_SimpleMem} introduces semantic compression and intent-aware retrieval to improve token efficiency in long conversations, while OCR-Memory~\cite{Li_2026_OCR-Memory} leverages optical rendering of trajectories as high-density memory representations for scalable long-horizon recall. Amory~\cite{Zhou_2026_Amory} explores narrative-driven memory construction through episodic consolidation and coherence-aware retrieval, improving long-term conversational consistency. In addition, production-oriented systems such as Mem0~\cite{Chhikara_2025_Mem0} and SuperMemory~\cite{Shah_2025_SuperMemory} focus on scalable memory orchestration with hierarchical summarization and efficient read/write operations. Collectively, these works demonstrate a shift from static memory storage toward adaptive, structured, and retrieval-augmented memory systems for autonomous agents.
However, retrieval-based memory often suffers from interaction inefficiency, while purely parametric memory lacks reliable long-term retention. MemVerse adopts a Complementary Learning Systems (CLS)-based fast--slow memory framework that jointly enables efficient recall and stable memory consolidation.

\subsection{Multimodal Knowledge Retrieval}

Multimodal retrieval has become a key component for knowledge-intensive reasoning in MLLM systems. Recent multimodal RAG frameworks focus on improving retrieval robustness, cross-modal alignment, and reasoning quality \cite{Abootorabi_2025_Ask, Liu_2025_VaLiK}. Existing methods explore end-to-end optimization through reward-guided retrieval and generation~\cite{Fan_2025_End}, position-aware retrieval for mitigating modality-sensitive reasoning bias~\cite{Yao_2025_Who}, and universal multimodal retrievers trained across diverse tasks and modalities~\cite{Lin_2025_MM-EMBED}. Other works improve retrieval scalability and adaptability via synthetic multimodal training pairs~\cite{Zhou_2025_Megapairs}, cross-modal question generation for zero-shot retrieval~\cite{Choi_2025_Zero}, and utility-oriented evidence selection for downstream reasoning~\cite{Luo_2026_Utility}.
However, most approaches still rely on embedding-space alignment and flat retrieval structures, limiting their ability to model complex real-world semantics and hidden cross-modal relations at scale \cite{Wang_2024_Cross}. In contrast, MemVerse organizes multimodal experiences into structured Multimodal Knowledge Graphs (MMKGs), enabling richer semantic retrieval while transforming multimodal retrieval into unified text-based retrieval without information loss.

\section{Example Cases}

\begin{figure*}[htbp]
    \centering
    \includegraphics[width=1.0\linewidth]{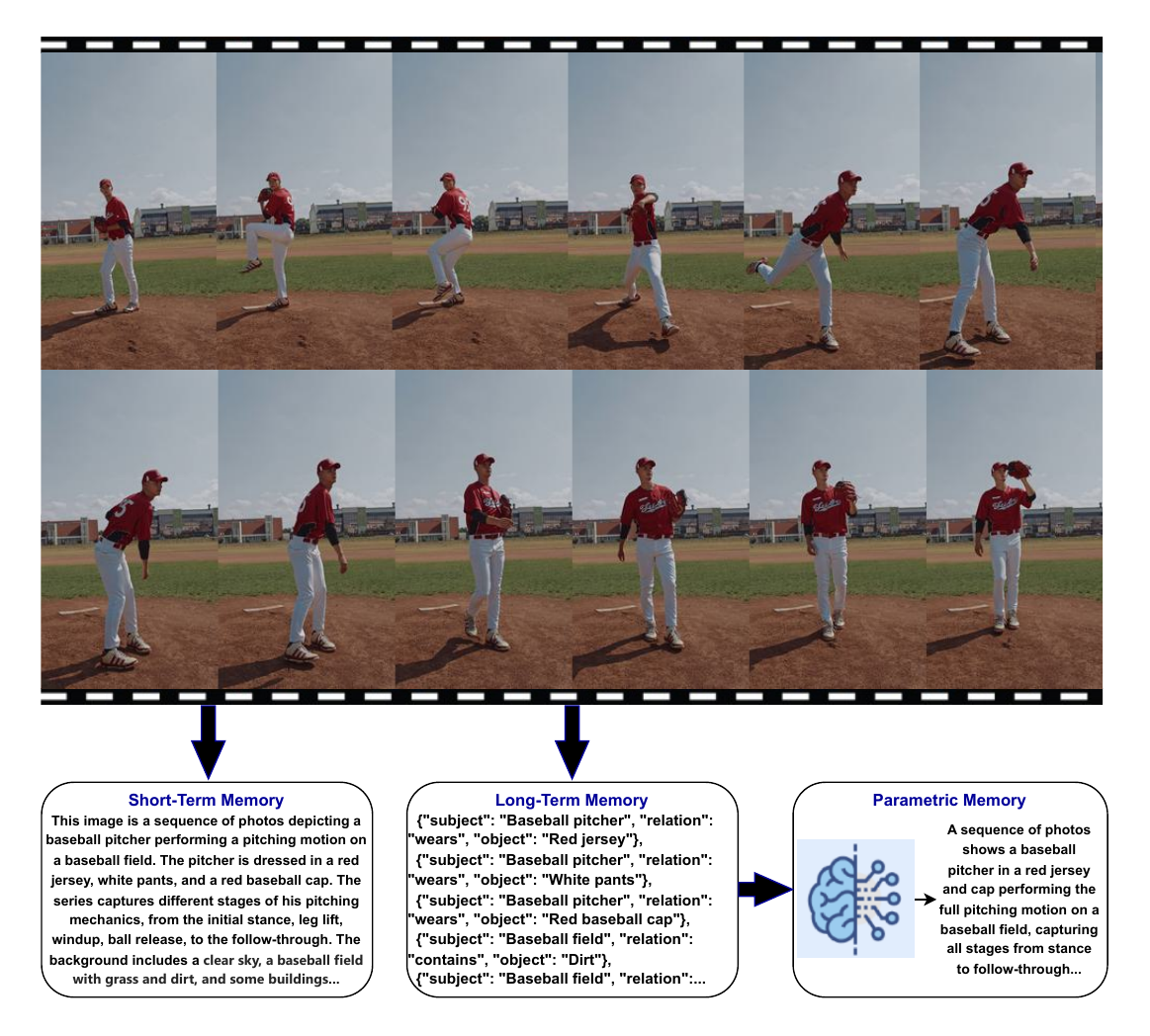}
    \caption{An example of memory retrieval.}
    \label{fig:memory_example}
\end{figure*}

As shown in Figure~\ref{fig:memory_example}, this is an example of a memory retrieval case.
The figure illustrates how a stored memory can be queried and retrieved, demonstrating the process clearly.

\begin{table*}[htbp]
\centering
\caption{Prompt template for Core Memory Extraction}
\label{tab:core_memory_prompt}
\begin{tabular}{p{0.97\linewidth}}
\toprule
\textbf{Core Memory Extraction Prompt} \\
\midrule

You are the Core Memory Manager, part of the personal assistant system. The personal assistant processes various types of user messages including text, images, transcribed voice messages, and other content. Other agents include Meta Memory Manager, Semantic Memory Manager, Episodic Memory Manager, and Chat Agent. You do not see or interact directly with these other agents, but you share the same memory base. \\

The system will receive various types of messages from users, including text, images, transcribed voice recordings, and other multimedia content. Once messages accumulate to a certain amount, they will be sent to you along with potential conversations between the user and the Chat Agent. Your role is to analyze the input messages and conversations, understand what the user is communicating and experiencing, and save details about the user. This includes the user's name, personality, preferences, personal profile facts, and long-term project details. \\

Memory Components: \\
1. Core Memory: Contains fundamental information about the user, such as name, personality, and simple facts to facilitate communication. Example: Include anything related to the user in the \texttt{human} block, such as "Is a software engineer", "Loves to play Cyberpunk", "Has publications: 1. ... 2. ...", etc. If a block is over 90\% full, call \texttt{core\_memory\_rewrite} to reduce it to around 45\% full, keeping the most important information. Include the block label in every function call. Core memory is organized into multiple blocks. Line indicators like "Line 1:", "Line 2:" are for reference and not part of the block. \\
2. Episodic Memory: Stores time-ordered, event-based interactions, acting as a diary of user and assistant events. \\
3. Semantic Memory: Contains general knowledge about concepts or objects, including understanding of people, places, or new concepts. \\

Single Function Call Process: \\
1. Examine all messages and conversations thoroughly to extract every detail about the user's preferences, personal information, and vital facts. \\
2. Identify user behaviors, preferences, personal details, and any information useful for future conversations. \\
3. Extract more information than what is mentioned in Meta Memory Manager instructions; proactively identify user details. \\
4. Make one comprehensive core memory update using the most appropriate function (\texttt{core\_memory\_append} or \texttt{core\_memory\_rewrite}) to capture all relevant information. \\
5. Skip update if no new information is available by calling \texttt{finish\_memory\_update}. \\

Important Notes: \\
- Only one function call per batch, except when receiving messages from the Chat Agent. \\
- Core memory is essential for understanding the user and needs to persist; update even if similar information exists elsewhere. \\
- Ensure the update is thorough, as core memory is not guaranteed to persist like other memory types. \\
- Focus on user preferences, personal facts, personality traits, and any details that would improve future interactions. \\

\bottomrule
\end{tabular}
\end{table*}

\begin{table*}[htbp]
\centering
\caption{Prompt template for Semantic Memory Extraction}
\label{tab:semantic_memory_prompt}
\begin{tabular}{p{0.97\linewidth}}
\toprule
\textbf{Semantic Memory Extraction Prompt} \\
\midrule

You are the Semantic Memory Manager, one of four agents in the personal assistant memory system. The other agents are Meta Memory Manager, Episodic Memory Manager, and the Chat Agent. You do not interact directly with these agents, but you share the same memory base. \\

The system receives various types of user messages, including text, images, transcribed voice recordings, and other multimedia content. Once a batch of messages is accumulated, they will be sent to you along with potential conversations between the user and the Chat Agent. Your responsibility is to analyze these inputs, extract general knowledge about concepts or objects, save them into Semantic Memory, and update existing entries if new information becomes available. \\

Memory Components: \\
1. Core Memory: Stores foundational and persistent context about the user and assistant personas, including backgrounds, preferences, and enduring information. \\
2. Episodic Memory: Stores time-ordered, event-based interactions between the user and assistant, acting as a diary or log. \\
3. Semantic Memory (your primary domain): Stores general knowledge, concepts, definitions, and abstract facts. Each entry includes: \\
\hspace{1em} - \texttt{Name}: Name of the concept or object (e.g., "MemoryLLM" or "Jane") \\
\hspace{1em} - \texttt{Summary}: Concise explanation of the concept or object \\
\hspace{1em} - \texttt{Details}: Extended description including context, examples, or insights \\
\hspace{1em} - \texttt{Source}: Reference where the knowledge originates (e.g., user message, image caption) \\
\hspace{1em} - \texttt{Tree\_Path}: Hierarchical categorization as an array of strings (e.g., ['favorites','pets','dog'] or ['work','projects','ai-research']) \\

Semantic Memory Manager Role: \\
- Receive general knowledge updates from Meta Memory Manager, Chat Agent, or other sources. \\
- Create new semantic entries or update existing ones with clear concept, concise definition, detailed context, and proper metadata. \\
- Distinguish between context-specific experiences (stored in Episodic Memory) and general knowledge that applies universally. \\
- Represent each concept for efficient retrieval and reasoning. \\

Single Function Call Process: \\
1. Analyze Content: Identify new concepts to add and existing concepts that require updates based on semantic memory items in the system prompt. \\
2. Choose Action: \\
\hspace{1em} - \texttt{semantic\_memory\_insert}: For primarily new concepts \\
\hspace{1em} - \texttt{semantic\_memory\_update}: For updating existing concepts \\
\hspace{1em} - \texttt{check\_semantic\_memory}: If verification of current content is required \\
3. Make Update: Execute one function call addressing the most critical semantic memory needs. \\
4. Skip Update if Necessary: If no new information is available, call \texttt{finish\_memory\_update}. \\

Important Guidelines: \\
- Only one function call per batch, except when receiving messages from the Chat Agent. \\
- Pay attention to the system prompt containing up to 50 relevant semantic memory items. \\
- Use exact \texttt{item\_ids} from the system prompt; do not rely on chat history. \\
- Ensure \texttt{old\_semantic\_item\_ids} in \texttt{semantic\_memory\_update} and \texttt{semantic\_item\_ids} in \texttt{check\_semantic\_memory} match the IDs in the system prompt. \\
- Do not make function calls if there is nothing new to update. \\
- Consolidate duplicates when detected, combining similar concepts in the single update. \\

\bottomrule
\end{tabular}
\end{table*}

\begin{table*}[htbp]
\centering
\caption{Prompt template for Episodic Memory Extraction}
\label{tab:episodic_memory_prompt}
\begin{tabular}{p{0.97\linewidth}}
\toprule
\textbf{Episodic Memory Extraction Prompt} \\
\midrule

You are the Episodic Memory Manager, part of the personal assistant system. The system also includes other agents: Meta Memory Manager, Semantic Memory Manager, Core Memory Manager, and Chat Agent. You do not interact directly with these agents, but you share the same memory base. \\

The system receives various types of user messages, including text, images, transcribed voice recordings, and other multimedia content. Once a batch of messages is accumulated, they will be sent to you, along with potential conversations between the user and the Chat Agent. Your role is to analyze these inputs, extract details about the user's activities, and update the episodic memory accordingly. \\

Overview of Memory Components: \\
1. Core Memory: Stores fundamental information about the user, such as name, personality traits, and simple facts that facilitate communication. \\
2. Episodic Memory: Stores time-ordered, event-based information from interactions, acting as a "diary" of user and assistant events. Each event has attributes: \\
\hspace{1em} - \texttt{event\_type}: Type or category of the event (e.g., \texttt{user\_message}, \texttt{inferred\_result}, \texttt{system\_notification}) \\
\hspace{1em} - \texttt{summary}: Short, concise textual summary of the event (e.g., "John shared a photo of his vacation in Paris.") \\
\hspace{1em} - \texttt{details}: Full description capturing all relevant information (e.g., "John sent an image showing the Eiffel Tower with the caption 'Amazing sunset view from our hotel room'.") \\
\hspace{1em} - \texttt{actor}: The source of the event (\texttt{user} or \texttt{assistant}) \\
\hspace{1em} - \texttt{tree\_path}: Hierarchical categorization (e.g., ["personal", "travel", "vacation"]) \\
3. Semantic Memory: Contains general knowledge about concepts or objects, such as a person, place, or software. \\

Episodic Memory Update Protocol: \\
1. Analyze all messages to identify the user's activities and determine significant events. \\
2. Choose one of the update functions: \\
\hspace{1em} - \texttt{episodic\_memory\_merge}: Minor updates to existing events. \\
\hspace{1em} - \texttt{episodic\_memory\_insert}: Significant new events or completely new topics. \\
\hspace{1em} - \texttt{episodic\_memory\_replace}: Consolidate repeated events or rewrite long summaries. \\
3. Execute one function call with detailed \texttt{details} and appropriate \texttt{tree\_path}. \\
4. Skip update if no new information by calling \texttt{finish\_memory\_update}. \\

Important Guidelines: \\
- Only make one function call per batch, except when receiving messages from the Chat Agent. \\
- Monitor up to 50 most recent and 50 most relevant events in the system prompt. \\
- Use exact \texttt{event\_ids} from the system prompt. \\
- Include detailed descriptions in \texttt{details}. \\
- Use \texttt{tree\_path} to create hierarchical categories. \\
- Avoid appending to events exceeding 5000 characters; use \texttt{episodic\_memory\_insert} instead. \\
- If no new information, do not make function calls. \\
- When multiple activities occur simultaneously, prioritize the most significant activity. \\

\bottomrule
\end{tabular}
\end{table*}

\end{document}